\documentclass[review]{elsarticle}

\usepackage{lineno,hyperref}
\modulolinenumbers[5]
\journal{Journal of Digital Communications and Networks}

\usepackage{array}

\usepackage[utf8]{inputenc}
\usepackage[ruled]{algorithm2e}
\usepackage[usenames,dvipsnames]{xcolor}
\usepackage{amsmath}
\usepackage{bbm}
\usepackage[noend]{algpseudocode}
\DeclareMathOperator*{\argmin}{arg\,min}
\DeclareMathOperator*{\argmax}{arg\,max}

\SetAlFnt{\small}
\SetAlCapFnt{\small}
\SetAlCapNameFnt{\small}
\SetAlCapHSkip{0pt}
\IncMargin{-\parindent}
\newcommand{\specialcell}[2][c]{
  \begin{tabular}[#1]{@{}c@{}}#2\end{tabular}}
\begin{document}

\begin{frontmatter}

\title{Machine Learning for Internet of Things Data Analysis: A Survey}

\author{Mohammad Saeid Mahdavinejad\fnref{mahdavi}}
\fntext[mahdavi]{University of Isfahan, Kno.e.sis - Wright State University}

\author{Mohammadreza Rezvan\fnref{rezvan} }
\fntext[rezvan]{University of Isfahan, Kno.e.sis - Wright State University}

\author{Mohammadamin Barekatain\fnref{barekatain}}
\fntext[barekatain]{Technische Universität München}

\author{Peyman Adibi\fnref{adibi}}
\fntext[adibi]{University of Isfahan}

\author{Payam Barnaghi\fnref{barnaghi}}
\fntext[barnaghi]{University of Surrey}
\cortext[mycorrespondingauthor]{Corresponding author}
\ead{p.barnaghi@surrey.ac.uk}

\author{Amit P. Sheth\fnref{sheth}}
\fntext[sheth]{ Kno.e.sis - Wright State University}




\begin{abstract}
Rapid developments in hardware, software, and communication technologies have allowed the emergence of Internet-connected sensory devices that provide observation and data measurement from the physical world.  By 2020, it is estimated that the total number of  Internet-connected devices being used will be between 25-50 billion. As the numbers grow and technologies become more mature, the volume of data published will increase. Internet-connected devices technology, referred to as Internet of Things (IoT), continues to extend the current Internet by providing connectivity and interaction between the physical and cyber worlds. In addition to increased volume, the IoT generates Big Data characterized by velocity in terms of time and location dependency, with a variety of multiple modalities and varying data quality. Intelligent processing and analysis of this Big Data is the key to developing smart IoT applications. This article assesses the different machine learning methods that deal with the challenges in IoT data by considering smart cities as the main use case. The key contribution of this study is presentation of a taxonomy of machine learning algorithms explaining how different techniques are applied to the data in order to extract higher level information. 
The potential and challenges of machine learning for IoT data analytics will also be discussed. 
A use case of applying Support Vector Machine (SVM) on Aarhus Smart City traffic data is presented for a more detailed exploration.
\end{abstract}

\end{frontmatter}


%
%


%
%




\section{Introduction}

Emerging technologies in recent years and major enhancements to Internet protocols and computing systems, have made the communication between different devices easier than ever before. According to various forecasts, around 25-50 billion devices are expected to be connected to the Internet by 2020. This has given rise to the newly developed concept of Internet of Things (IoT). IoT is a combination of embedded technologies regarding wired and wireless communications, sensor and actuator devices, and the physical objects connected to the Internet \cite{cite61,cite3}. One of the long-standing objectives of computing is to simplify and enrich human activities and experiences (e.g., see the visions associated with “The Computer for the 21st Century” \cite{weiser1999c} or “Computing for Human Experience” \cite{sheth2010c}) IoT needs data to  either represent better services to users or enhance IoT framework performance to accomplish this intelligently. In this manner, systems should be able to access raw data from different resources over the network and analyze this information to extract knowledge.

Since IoT will be among the greatest sources of new data, data science will make a great contribution to make IoT applications more intelligent. Data science is the combination of different fields of sciences that uses data mining, machine learning and other techniques to find patterns and new insights from data. These techniques include a broad range of algorithms applicable in different domains. 
The process of applying data analytics methods to particular areas involves defining data types such as volume, variety, velocity; data models such as neural networks, classification, clustering methods and applying efficient algorithms that match with the data characteristics. By following our reviews, it is deduced that: firstly, since data is generated from different sources with specific data types, it is important to adopt or develop algorithms that can handle the data characteristics, secondly, the great number of resources that generate data in real time are not without the problem of scale and velocity and thirdly, finding the best data model that fits the data is one of the most important issues for pattern recognition and for better analysis of IoT data.
These issues have opened a vast number of opportunities in expanding new developments. Big Data is defined as high-volume, high-velocity, and high variety data that demand cost-effective, innovative forms of information processing which enable enhanced insight, decision making, and process automation\cite{cite68}.

With respect to the challenges posed by Big Data, it is necessary to divert to a new concept termed \textit{Smart Data}, which means: "realizing productivity, efficiency, and effectiveness gains by using semantics to transform raw data into Smart Data" \cite{cite69} . A more recent definition of this concept is: "Smart Data provides value from harnessing the challenges posed by volume, velocity, variety, and veracity of Big Data, and in turn providing actionable information and improving decision making." \cite{cite691}. At last, Smart Data can be a good representative for IoT data. 
\subsection{The Contribution of this paper}

The objective here is to answer the following questions:

\textit{A})\textbf{How could machine learning algorithms be applied to IoT smart data?}

\textit{B})\textbf {What is the taxonomy of machine learning algorithms that can be adopted in IoT?}

\textit{C})\textbf {What are IoT data characteristics in real-world?}

\textit{D})\textbf {Why is the Smart City  a typical use case of IoT applications?}

A) To understand which algorithm is more appropriate for processing and decision-making on generated smart data from the things in IoT, realizing these three concepts is essential. First, the IoT application (Sec. \ref{IoT}), second, the IoT data characteristics (Sec, \ref{Smart-C-data}), and the third, the data-driven vision of machine learning algorithms (Sec. \ref{ML}). We finally discussed the issues in Sec. \ref{Disc}.

B) About 70 articles in the field of IoT data analysis are reviewed, revealing that there exist eight major groups of algorithms applicable to IoT data. These algorithms are categorized according to their structural similarities, type of data they can handle, and the amount of data they can process in reasonable time.

C) Having reviewed the real-work perspective of  how IoT data is analyzed by over 20 authors, many significant and insightful results have been revealed regarding data characteristics. We discussed the results in Sec. \ref{Disc} and Table \ref{tab:four}. To have a deeper insight into IoT smart data, patterns must be extracted and the generated data interpreted. Cognitive algorithms will undertake interpretation and matching, much as the human mind would do.  Cognitive IoT systems will learn from the data previously generated and will improve when performing repeated tasks. Cognitive computing as as a prosthetic for human cognition by analyzing massive amounts of data and being able to respond to questions humans might have when making certain decisions. Cognitive Iot plays an important role in enabling the extraction of meaningful patterns form the IoT smart data generated \cite{cite71}.

D) Smart City has been selected as our primary use case in IoT for three reasons:
Firstly, among all of the reviewed articles the focus of 60 percents is on the field of the Smart City, secondly, Smart City includes many of the other use cases in IoT, and thirdly, there are many open datasets for Smart City applications easily accessible for researchers.
Also, Support Vector Machine (SVM) algorithm is implemented on the Aarhus City smart traffic data in order to predict traffic hours during one day in Sec. \ref{Disc}.
By answering the above questions about the IoT smart data and machine learning algorithms, we would be able to choose the best machine learning algorithm that can handle IoT smart data characteristics. Unlike the others, similar surveys about the machine learning and IoT, readers of this article would be able to get deep and technical understanding  of machine learning algorithms, IoT applications, and IoT data characteristics along with both technical and simple implementations.

\subsection{Organization} The rest of this paper is organized as follows:
 the related articles in this field are reviewed and reported in Sec. \ref{RW}. IoT applications and communication protocols, computing frameworks, IoT architecture, and Smart City segments are reviewed, explained, briefed and illustrated in Sec. \ref{IoT}. The quality of data, Big Data generation, integrating sensor data and semantic data annotation are reviewed in Sec. \ref{SC}. Machine learning algorithms in eight categories based on recent researches on IoT data and frequency of machine learning algorithms are reviewed and briefed in Sec. \ref{ML}. Matching the algorithms to the particular Smart City applications is done in Sec. \ref{Disc}, and the conclusion together with future research trends and open issues are presented in Sec. \ref{Research-trends}.

\begin{figure}
\centerline{\includegraphics[width=1\linewidth]{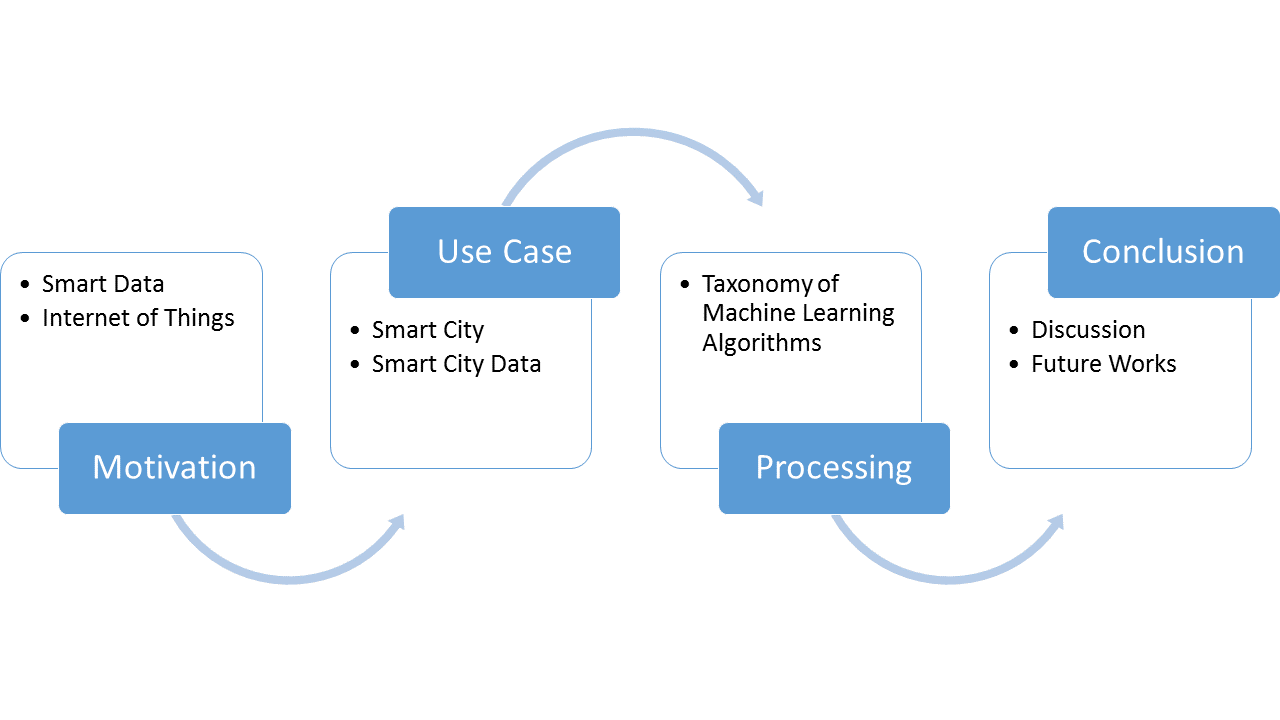}}

\caption{Organization of survey}
\label{fig:three}
\end{figure}

\section{Literature Review}
\label{RW}
Since  IoT represents a new concept for the Internet and smart data, it is a challenging area in the field of computer science. The important challenges for researchers with respect to  IoT consist of  preparing and processing data. 

\cite{cite29} proposed 4 data mining models for processing IoT data. 
The first proposed model is a \textit{multi layer model}, based on a data collection layer, a data management layer, an event processing model, and data mining service layer. The second model is a \textit{distributed data mining model}, proposed for data deposition at different sites. The third model is a \textit{grid based data mining model} where the authors seek to implement  heterogeneous, large scale and high performance applications, and the last model is a \textit{data mining model from multi technology integration perspective}, where the corresponding framework for a future Internet is described.

\cite{cite30} performed research into warehousing radio frequency identification, (RFID) data, with a focus on managing and mining RFID stream data, specifically.

\cite{cite31} introduce a systematic manner for reviewing data mining knowledge and techniques in most common applications. In this study, they reviewed some data mining functions like classification, clustering, association analysis, time series analysis, and outline detection. They revealed that the data generated by data mining applications such as e-commerce, Industry, healthcare, and city governance are similar to that of the IoT data. Following their findings, they assigned the most popular data mining functionality to the application and determined which data mining functionality was the most appropriate for processing each specific application’s data.

\cite{cite32} ran a survey to respond to some of the challenges in preparing and processing data on the IoT through data mining techniques. They divided their research into three major sections, in the first and second sections; they explain  IoT,  the  data, and the challenges that exist in this area, such as building a model of mining and mining algorithms for IoT. In the third section, they discuss the potential and open issues that exist in this field. Then, data mining on IoT data have  three major concerns: first, it must be shown that processing data will solve the chosen problems. Next the data characteristics must be extracted from generated data, and then, the appropriate algorithm is chosen according to the taxonomy of algorithms and data characteristics.

\cite{cite70}  attempted to explain the Smart City infrastructure in IoT and discussed the advanced communication  to support added-value services for the administration of the city and citizens thereof. They provide a comprehensive view of enabling technologies, protocols, and architectures for Smart City. In the technical part of their, the article authors reviewed the data of Padova Smart City.


\section{Internet of Things}
\label{IoT}
The purpose of  \textit{Internet of Things}, (IoT)  is to develop a smarter environment, and a simplified  life-style by saving time, energy, and  money. Through this technology, the expenses in different industries can be reduced.
The enormous investments and many studies running on IoT has made IoT a growing trend in recent years.
 IoT is a set of connected devices that can transfer data among one another in order to optimize their performance;
 these actions occur automatically and without human awareness or input.
 IoT includes four main components: 1) sensors, 2)processing networks, 3) analyzing data, and 4) monitoring the system.
The most recent advances made in IoT began when radio frequency identification (RFID) tags were put into use more frequently, lower cost sensors became more  available, web technology developed, and communication protocols changed \cite{cite4,cite5}. 
The IoT is integrated with different technologies and connectivity is necessary and  sufficient condition for
it. So communication protocols are constituents the technology that should be enhanced \cite{cite2,cite6}. 
In IoT, communication protocols can be divided into three major components:

(1)\textit    { Device to Device (D2D)}: this type of communication enables communication between nearby mobile phones. This is the next generation of cellular networks.

(2) \textit    { Device to Server (D2S)}: in this type of communication devices, all the data is sent to the servers, which can be close or far from the devices. This type of communication mostly is applied in cloud processing.

(3) \textit    { Server to Server (S2S)}:  in this type of communication, servers transmit data between each other. This type of communication mostly is applied in cellular networks.

Processing and preparing data for these communications is a critical challenge. To respond to this challenge, different kinds of data processing, such as analytics at the edge, stream analysis, and IoT analysis at the database, must be applied. 
The decision to apply any one of the mentioned processes depends on the particular application and its needs\cite{cite7}.
Fog and cloud processing are two analytical methods adopted in processing and preparing data before transferring to the other things.  The whole task of IoT is summarized as follows: first, sensors and IoT devices collect the information from the environment. Next, knowledge should be extracted from the raw data. Then, the data will be ready for transfer to  other objects, devices, or servers through the Internet.

\subsection{Computing Framework}

Another important part of IoT is \textit{computing frameworks}  for processing data, the most famous of which are fog and cloud computing.
IoT applications use both frameworks depending on application and process location. In some applications, data should be processed upon generation, while in other applications, it is not necessary to process data immediately. The instant processing of data and the network architecture that supports it is known as fog computing. Collectively, they are applied for edge computing\cite{cite11}.

\subsubsection{\textit	{Fog Computing:}} Here, the architecture of fog computing is applied to migrate information from the data centers task to the edge of the servers. This architecture is built based on the edge servers. Fog computing provides limited computing, storage, and network services, also providing logical intelligence and filtering of  data for data centers.
This architecture has been and is being implemented in vital areas like eHealth and military applications \cite{cite12,cite13}.

\subsubsection{\textit	{Edge Computing:}} In this architecture, processing is run at a distance from the core, toward the edge of the network. This type of processing enables data to be initially processed at the edge devices. Devices at the edge may not be  connected to the network in a continuous manner, so they need a copy of master data/reference data for offline  processing. Edge devices have different features such as
1)enhancing security, 
2)filtering and cleaning of the data, and 
3)storing local data for local use\cite{cite14}.

\subsubsection{\textit	{Cloud Computing:}} Here, data for processing is sent to the data centers, and  after being analyzed and processed, they become accessible.

This architecture has high latency and high load balancing, indicating that this architecture is not sufficient enough for processing IoT data because most processing should run at high speeds. The volume of this data is high, and Big Data processing will increase the CPU usage of the cloud servers\cite{cite15}. There are different types of cloud computing:

(1)\textit { Infrastructure as a Service (IaaS)}: where the company purchases all the equipment like hardware, servers , and networks.

(2)\textit { Platform as a Service (PaaS)}: where all the equipment above, are put for rent on the Internet.

(3)\textit { Software as a service(SaaS)}: where a distributed software model is presented. In this model, all the practical software will be hosted from a service provider, and practical software can be accessible to the users through the Internet \cite{cite16}.

(4)\textit { Mobile Backend as a Service (MBaaS)}: also known as a Backend as a Service(BaaS), provides the web and mobile application with a path in order to connect the application to the backend cloud storage. MBaaS provides features like user management, push notification and integrates with the social network services.
This cloud service benefits from application programming interface (API) and software development kits (SDK).

\subsubsection{\textit	{Distributed Computing}}:
This architecture is designed for processing high volume data. In IoT applications, because the sensors generate data on a repeated manner, Big Data challenges are encountered\cite{cite14,cite17}.
To overcome this phenomenon, a distributed computing is designed to divide data into packets, and assign each packet to different computers for processing.  This distributed computing has different frameworks like Hadoop and Spark.
 When migrating from cloud to fog and distributed computing, the following occur: 
1) a decrease in network loading,
2) an increase in data processing speed, 
3) a reduction in CPU usage, 
4) a reduction in energy consumption, and
5) higher data volume processing.

Because the Smart City is one of the primary applications of IoT, the most important use cases of Smart City and their data characteristics are discussed in the following sections.

\section{Smart City}
\label{SC}
Cities always demand services to enhance the quality of life and make services more efficient. In the last few years, the concept of smart cities has played an important role in academia and in industry  \cite{cite22}. With an increase in the population and complexity of city infrastructures,  cities seek manners to handle large-scale urbanization problems. IoT plays a vital role in collecting data from the city environment. IoT enables cities to use live status reports and smart monitoring systems to react more intelligently against the emerging situations such as earthquake and volcano. By using  IoT technologies in cities, the majority of the city’s assets can be connected to one another, make them more readily observable, and consequently, more easy to monitor and manage. The purpose of building smart cities is to improve services like traffic management, water management,  and energy consumption, as well as improving the quality of life for the citizens. The objectives of  smart cities is to transform rural and urban areas into places of democratic innovation \cite{cite24}. Such smart cities seek to decrease the expenses in public health, safety, transportation and resource management, thus assisting the their economy \cite{cite33}.

\cite{cite23} Believe that in the long term, the vision for a  Smart City would be  that all the cities' systems and structures will monitor their own conditions and carry out self-repair upon need.

\subsection{Use Case}
A city has an important effect on society because the city touches all aspects of human life. Having a Smart City can assist in having a comfortable life. Smart Cities use cases consist of Smart Energy, smart mobility, Smart Citizens, and urban planning.
This division is based on reviewing the latest studies in this field and the most recent reports released by McKinsey and Company.
\subsubsection{\textit{Smart Energy}}
Smart Energy is one of the most important research areas of IoT because it is essential to reduce overall power consumption\cite{cite27}.
It offers high-quality, affordable environment energy  friendly.
Smart Energy includes a variety of operational and energy measures, including Smart Energy applications, smart leak monitoring, renewable energy resources, etc.
Using Smart Energy(i.e., deployment of a smart grid) implies a fundamental re-engineering of the electricity services\cite{cite28}.
\textit{Smart Grid}  is one of the most important applications of Smart Energy. It includes many high-speed time series data to monitor key devices.  For managing this kind of data,  \cite{cite62}  have introduced a method to manage and analyze time series data in order to make them organized on demand. Moreover, Smart Energy infrastructure will become more complex in future, therefore \cite{cite63} have proposed a simulation system to test new concept and optimization approaches and forecast future consumption. Another important application of Smart Energy is a leak monitoring system. The objective of this system is to model a water or gas management system which would optimize energy resource consumption \cite{cite64,cite65}.

\subsubsection{\textit{Smart Mobility}}
Mobility is another important part of the city. Through the IoT, city officials can improve the quality of life in the city. Smart mobility can be divided into the following three major parts:

(1)\textit{ Autonomous cars}: IoT will have a broad range effect on how vehicles are run. The most important question is about how IoT can improve vehicle services. IoT sensors and wireless connections make it possible to create self-driving cars and monitor vehicles  performance. With the data collected from vehicles, the most popular/congested routes can be predicted, and decisions can be made to decrease the traffic congestion. Self-driving cars can improve the passenger safely because they have the ability to monitor the driving of other cars.

(2)\textit{ Traffic control}: Optimizing the traffic flow by analyzing sensor data is another part of mobility in the city. In traffic control, traffic data will be collected from the cars, road cameras,  and from counter sensors installed on roads.

(3)\textit{ Public transportation}:  IoT can improve the public transportation system management by providing accurate location and routing information to smart transportation system. It can assist the passengers in making better decisions in their schedules as well as decrease the amount of wasted time.
There exist different perspectives over how to build smart public transportation systems. These systems need to manage a different kind of data like vehicle location data and traffic data. Smart public transportation systems should be real-time oriented in order to make proper decisions in real-time as well as use historical data analysis \cite{cite66}. For instance, \cite{cite67} have proposed a mechanism that considers Smart City devices as graph nodes and they have used Big Data solutions to solve these issues. 

\subsubsection{\textit{Smart Citizen}}
This use case for Smart Cities covers a broad range of areas in human lives, like environment monitoring, crime monitoring, and social health. The environment with all its components is fundamental and vital for life; consequently, making  progress in technology is guaranteed to enhance security. Close monitoring devoted to crime would also contribute to overall social health .

\subsubsection{\textit{Urban Planning}}
Another important aspect in use cases for the Smart City is drawing long-term decisions. Since the city and environment have two major roles in human life, drawing decisions in this context is critical.
By collecting data from different sources, it is possible to draw a decision for the future of the city. Drawing decisions affecting the city infrastructure, design, and functionality is called urban planning. IoT is beneficial in this area because through Smart City data analysis, the authorities can predict which part of the city will be more crowded in the future and find solutions for the potential problems. A combination of  IoT and urban planning would have a major affect on scheduling future infrastructure improvements.

\subsection{Smart City Data Characteristics }
\label{Smart-C-data}

Smart cities' devices generate data on a continuous manner, indicating that the data gathered from traffic, health, and energy management applications would provide sizable volume. In addition, since the data generation rate varies for different devices, processing data with different generation rates is a challenge. For example, frequency of GPS sensors updating is measured in seconds while, the frequency of updates for temperature sensors may be measured hourly. Whether the data generation rate is high or low, there always exists the danger of important information loss. To integrate the sensory data collected from heterogeneous sources is challenging\cite{cite4,cite40}.
\cite{cite37} applied Big Data analytic methods to distinguish the correlation between the temperature and traffic data in Santander City, Spain.

\cite{cite38} proposed a new framework integrating Big Data analysis and Industrial Internet of Things (IIoT) technologies for Offshore Support Vessels (OSV) based on a hybrid CPU/GPU high-performance computing platform.

Another characteristic is the dynamic nature of the data.  Autonomous cars’ data is an example of Dynamic Data because the sensor results will change based on different locations and times. 

The quality of the collected data is important, particularly the Smart City data which have different qualities due to the fact that they are generated from heterogeneous sources.
According to\cite{cite39}, the quality of information from each data source depends on three factors:

1) Error in measurements or precision of data collection.

2) Devices' noise in the environment.

3) Discrete observation and measurements.

\begin{figure}
\centerline{\includegraphics[width=1.2\linewidth]{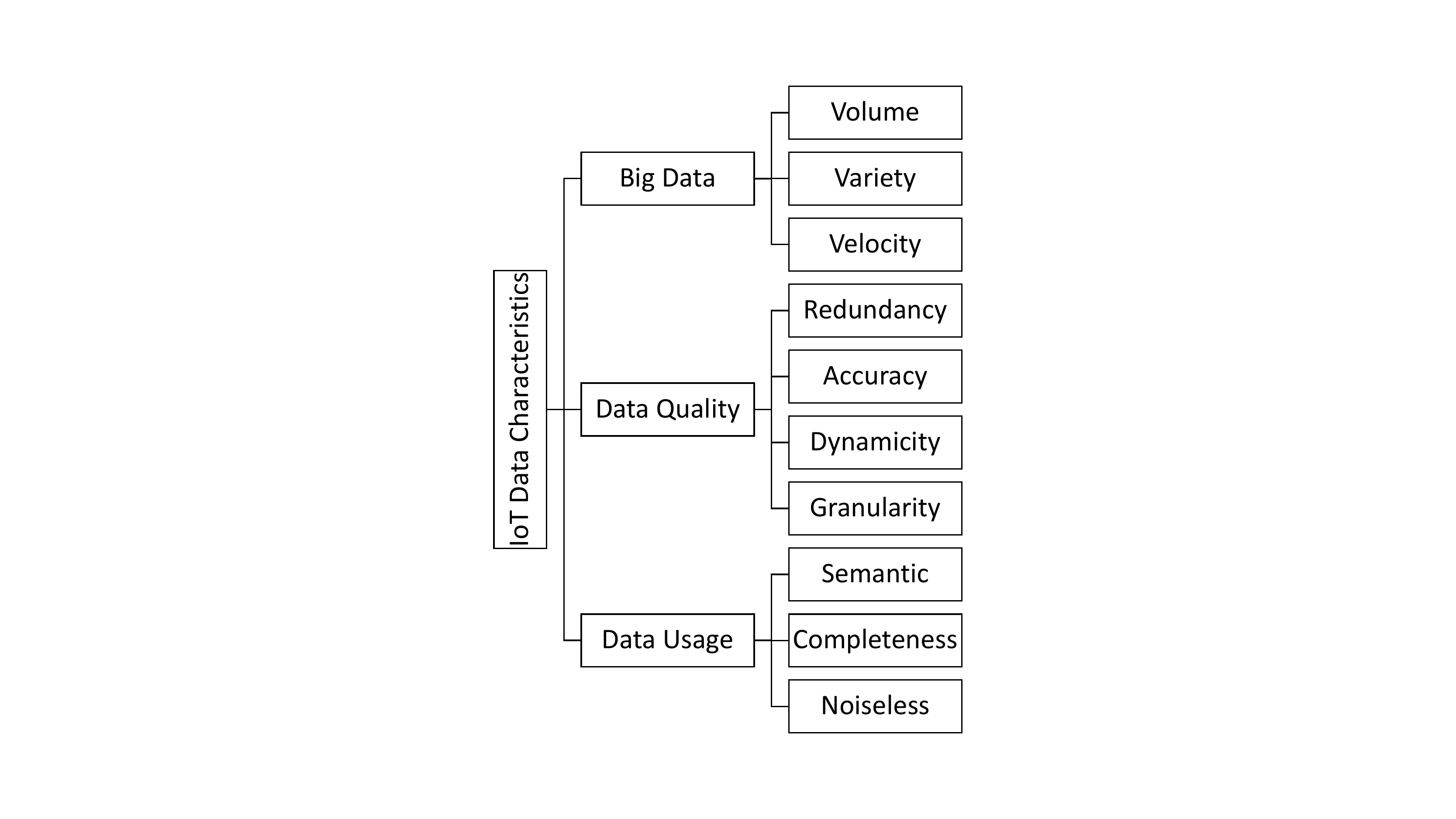}}

\caption{Data characteristics}
\label{fig:two}
\end{figure}

To have a better Quality of Information (QoI), it is necessary to extract higher levels of abstraction and provide actionable information to other services. QoI in Smart Data depends on the applications and characteristics of data. There exist different solutions to improve QoI. For example, to improve the accuracy of data, selecting  trustworthy sources and combining the data from multiple resources is of essence. By increasing frequency and density of sampling, precision of the observations and measurements will be improved, which  would lead a reduction in  environmental noise.
The data characteristics in both IoT and Smart City are shown in Figure \ref{fig:two}.
Semantic data annotation is another prime solution to enhance data quality. Smart devices generate raw data with low-level abstractions; for this reason, the semantic models will provide interpretable descriptions on data, its quality,  and original attributes \cite{cite33}. Semantic annotation is beneficial in interpretable and knowledge-based information fusion\cite{cite36}.
Smart data characteristics in smart cities are tabulated in brief, in Table \ref{tab:four}.

\newcolumntype{C}[1]{>{\centering\arraybackslash}p{#1}}

\begin{table}[htbp]
   \centering
   \caption{Characteristic of Smart Data in smart cities}
   \label{tab:four}
\begin{tabular}{|C{3.5cm}|C{4cm}|C{3cm}|C{2cm}|}
\hline
\textbf{\specialcell{Smart City\\ Use Cases} } & \textbf{\specialcell{Type of Data}}& \textbf{\specialcell{Where Data\\ Processed}} &  \textbf{References}\\\hline

 Smart Traffic & Stream/Massive Data & Edge &\specialcell{\cite{cite43}\\ \cite{cite4}} \\\hline
 Smart Health & Stream/Massive Data & Edge/Cloud & \specialcell{\cite{cite49}}\\\hline
 Smart Environment & Stream/Massive Data & Cloud & \specialcell{\cite{cite58}}\\\hline
 Smart Weather Prediction & Stream Data & Edge & \specialcell{\cite{cite41}}\\\hline
 Smart Citizen &Stream Data & Cloud &\specialcell{\cite{cite55}\\ \cite{cite45}} \\\hline
 Smart Agriculture &Stream Data & Edge/Cloud & \specialcell{\cite{cite50}}\\\hline
 Smart Home &Massive/Historical Data & Cloud & \specialcell{\cite{cite46}}\\\hline
 Smart Air Controlling &Massive/Historical Data & Cloud & \specialcell{\cite{cite40}}\\\hline
 Smart Public Place Monitoring & Historical Data & Cloud & \specialcell{\cite{cite51}}\\\hline
 Smart Human Activity Control & Stream/Historical Data & Edge/Cloud &\specialcell{\cite{cite47} \\ \cite{cite52}} \\\hline

\end{tabular}
\end{table}

\section{Taxonomy of machine learning algorithms}
\label{ML}
Machine learning is a sub field of computer science, a type of Artificial Intelligence, (AI), that provides machines with the ability to learn without explicit programming. Machine learning evolved from pattern recognition and Computational Learning Theory. There, some essential concepts of machine learning are discussed as well as, the frequently applied machine learning algorithms for smart data analysis.

A learning algorithm takes a set of samples as an input named  a \textit{training set}. In general, there exist three main categories of learning: \textit{supervised}, \textit{unsupervised}, and \textit{reinforcement} \cite{barber2012,bishop1,murphy2012}. In an informal sense, in supervised learning, the training set consists of samples of input vectors together with their corresponding appropriate target vectors, also known as \textit{labels}. In  unsupervised learning, no labels are required for the training set. Reinforcement learning deals with the problem of learning the appropriate action or sequence of actions to be taken for a given situation in order to maximize payoff. This article focuses is on supervised and unsupervised learning since they have been and are being widely applied in IoT smart data analysis.
The objective of supervised learning is to learn how to predict the appropriate output vector for a given input vector. Applications where the target label is a finite number of discrete categories are known as \textit{classification} tasks. Cases where the target label is composed of one or more continuous variables are known as \textit{regression} \cite{Goodfellow-et-al-2016}.\\
Defining the objective of unsupervised learning is difficult. One of the major objectives is to identify the sensible clusters of similar samples within the input data, known as \textit{clustering}. Moreover, the objective may be the discovery of a useful internal representation for the input data by \textit{preprocessing} the original input variable in order to transfer it  into a new variable space. This preprocessing stage can significantly improve the result of the subsequent machine learning algorithm and is named \textit{feature extraction} \cite{bishop1}.

The frequently applied machine learning algorithms for smart data analysis are tabulated in Table \ref{tab:one}.

In the following subsections, we assume that we are given a training set containing $N$ training samples denoted as $\{(x_i,y_i)\}_{i=1}^N$, where $x_i$ is the the $i$\textsuperscript{th} training $M$-dimensional input vector and $y_i$ is it's corresponding desired $P$-dimensional output vector. Moreover, we collect the \textit {M}-dimensional input vectors into a matrix, written $\textbf{x}\equiv(x_1,\ldots,x_N)^T$, and  we also collect their corresponding desired output vectors in a matrix, written $\textbf{y}\equiv(y_1,\ldots,y_N)^T$. However, in Section \ref{Cluster} the training set does not contain the desired output vectors.

\begin{table}[htbp]
   \centering
   \caption{Overview of frequently used machine learning algorithms for smart data analysis}
   \label{tab:one}
\begin{tabular}{|C{3.85cm}|C{4cm}|C{1.7cm}|C{3cm}|}
\hline
\textbf{Machine learning algorithm} & \textbf{Data processing tasks} & \textbf{Section} & \textbf{\specialcell{Representative\\references}}\\\hline
K-Nearest Neighbors & Classification & \ref{KNN} & \specialcell{\cite{cover1967} \cite{jagadish2005}}\\\hline
Naive Bayes & Classification & \ref{NaiB} & \specialcell{\cite{mccallum1998}  \cite{zhang2004o}}\\\hline
Support Vector Machine & Classification & \ref{SVM} & \specialcell{\cite{cortes1995s} \cite{guyon1993a}  \cite{cristianini2000i} \\ \cite{scholkopf2001l}}\\\hline
Linear Regression & Regression & \ref{LinReg} & \specialcell{\cite{neter1996} \cite{neter1996} \cite{seber2012} \\ \cite{montgomery2015}} \\\hline
Support Vector Regression & Regression & \ref{RVM} & \specialcell{\cite{smola1997s}\\ \cite{smola2004t}}\\\hline
Classification and Regression Trees & Classification/Regression & \ref{Tree} & \specialcell{\cite{breiman1984} \cite{prasad2006} \\ \cite{loh2011} }\\\hline
Random Forests & Classification/Regression & \ref{RandF} & \cite{breiman2001}\\\hline
Bagging & Classification/Regression & \ref{Bagg} & \cite{breiman1996b}\\\hline
K-Means & Clustering & \ref{KMeans} &  \specialcell{\cite{likas2003}  \cite{Ng1}  \cite{jumutc2015}}\\\hline
\specialcell{Density-Based Spatial \\Clustering of Applications\\ with Noise} & Clustering & \ref{DBSCAN} &  \specialcell{\cite{ester1996d} \\ \cite{kriegel2011d} \\ \cite{campello2013d}}\\\hline
Principal Component Analysis & Feature extraction & \ref{PCA} & \specialcell{\cite{pearson1901} \cite{hotelling1933}  \cite{jolliffe2002}  \cite{abdi2010}\\ \cite{bro2014}}\\\hline
Canonical Correlation Analysis & Feature extraction & \ref{CCA} &  \specialcell{\cite{hotelling1936}\\ \cite{bach2002}}\\\hline
Feed Forward Neural Network & \specialcell{Regression/Classification/\\Clustering/Feature extraction} & \ref{FFNN}  & \specialcell{\cite{lecun1998} \cite{glorot2010understanding}  \cite{eberhart2014}  \cite{he2015} \cite{lecun2015} \\ \cite{Goodfellow-et-al-2016}} \\\hline
One-class Support Vector Machines & Anomaly detection & \ref{OCSVM} & \specialcell{\cite{scholkopf2001e} \\ \cite{ratsch2002c}}\\\hline
\end{tabular}
\end{table}

\subsection{Classification}
\subsubsection{K-Nearest Neighbors}
\label{KNN}
In K-nearest neighbors (“KNN”), the objective is to classify a given new, unseen data point by looking at $K$ given data points in the training set, which are closest in input or feature space. Therefore, in order to find the $K$ nearest neighbors of the new data point, we have to use a distance metric such as Euclidean distance, $L_\infty$ norm, angle, Mahalanobis distance or Hamming distance. To formulate the problem, let us denote the new input vector (data point) by $x$, it’s $K$ nearest neighbors by $N_k(x)$, the predicted class label for $x$ by $y$, and the class variable by a discrete random variable $t$. Additionally, $\mathbbm{1}{(.)}$ denotes \textit{indicator function}: $\mathbbm{1}{(s)} = 1$ if $s$ is true and $\mathbbm{1}{(s)} = 0$ otherwise. The form of the classification task is
 
\begin{equation}
\label{eqn:03}
\begin{split}
&p (t = c | x, K) = \frac{1}{K} \sum_{i \in N_k(x)} \mathbbm{1}{(t_i = c)}, 
\\ & y = \argmax_cp(t = c | x, K)
\end{split}
\end{equation}
i.e., the input vector $x$ will be labeled by the mode of its neighbors’ labels \cite{cover1967}.\\

One limitation of KNN is that it requires storing the entire training set, which makes KNN unable to scale large data sets. In \cite{jagadish2005}, authors have addressed this issue by constructing a tree-based search with some one-off computation. Moreover, there exists an online version of KNN calcification. It is worth noting that KNN can  also be used for regression task \cite{bishop1}. However we don’t explain it here, since it is not a frequently used algorithm for smart data analysis.
\cite{cite56}  proposes a new framework for learning a combination of multiple metrics for a robust KNN classifier. Also, \cite{cite55} compares K-Nearest Neighbor with a rough-set-based algorithm for classifying the travel pattern regularities.

\subsubsection{Naive Bayes}
\label{NaiB}
Given a new, unseen data point (input vector) $z = (z_1,\ldots,z_M)$, naive Bayes classifiers, which are a family of probabilistic classifiers, classify $z$ based on applying Bayes’ theorem with the “naive” assumption of independence between the features (attributes) of $z$ given the class variable $t$. By applying the Bayes' theorem we have
\begin{equation}
\label{eqn:nb}
p (t = c | z_1,\ldots,z_M) = \frac{p(z_1,\ldots,z_M | t = c)p(t = c)}{p(z_1,\ldots,z_M)}
\end{equation}
and by applying the naive independence assumption and some simplifications we have
\begin{equation}
\label{eqn:nb2}
p (t = c | z_1,\ldots,z_M) \propto p(t = c)\prod_{j=1}^M p(z_j|t=c)
\end{equation}
Therefore, the form of the classification task is
\begin{equation}
\label{eqn:nb3}
y = \argmax_c{p(t = c)\prod_{j=1}^M p(z_j|t=c)}
\end{equation}
where $y$ denotes the predicted class label for $z$. The different naive Bayes classifiers use different approaches and distributions to estimate $p(t=c)$ and $p(z_j|t=c)$ \cite{zhang2004o}.\\

Naive Bayes classifiers require a small number of data points to be trained, can deal with high-dimensional data points, and are fast and highly scalable \cite{mccallum1998}. Moreover, they are a popular model for applications such as spam filtering \cite{metsis2006}, text categorization, and automatic medical diagnosis \cite{webb2005}.
\cite{cite50} used this algorithm to combine factors to evaluate the trust value and calculate the final quantitative trust of the Agricultural product. 
\subsubsection{Support Vector Machine}
\label{SVM}
The classical Support Vector Machines (SVMs) are non-probabilistic, binary classifiers that aim at finding the dividing hyperplane which separates both classes of the training set with the maximum margin. Then, the predicted label of a new, unseen data point, is determined based on which side of the hyperplane it falls  \cite{cortes1995s}. First, we discuss the Linear SVM that finds a hyperplane, which is a linear function of the input variable. To formulate the problem, we denote the normal vector to the hyperplane by $w$ and the parameter for controlling the offset of the hyperplane from the origin along its normal vector by $b$. Moreover, in order to ensure that SVMs can deal with outliers in the data, we introduce variable $\xi_i$, that is, a \textit{slack variable}, for every training point $x_i$ that gives the distance of how far this training point violates the margin in the units of $|w|$. This binary linear classification task is described using a constrained optimization problem of the form  
\begin{equation}
\label{eqn:SVM}
\begin{aligned}
& \underset{w, b, \xi}{\text{minimize}}
& &f(w,b,\xi) = \frac{1}{2}w^Tw + C\sum^n_{i=1} \xi_i \\
& \text{subject to}
& & y_i(w^Tx_i+b)-1+\xi_i\geq 0 & i = 1,\ldots,n,\\
&
& & \xi_i\geq 0 &i = 1,\ldots,n.
\end{aligned}
\end{equation}
where parameter $C>0$ determines how heavily a violation is punished \cite{scholkopf2001l,guyon1993a}. It should be noted that although here we used $L_1$ norm for the penalty term $\sum^n_{i=1}\xi_i$, there exist other penalty terms such as $L_2$ norm which should be chosen with respect to the needs of the application. Moreover, parameter $C$ is a hyperparameter which can be chosen via cross-validation or Bayesian optimization.
To solve the constrained optimization problem of equation \ref{eqn:SVM}, there are various techniques such as quadratic programming optimization \cite{gould2000}, sequential minimal optimization \cite{platt1998s}, and P-packSVM \cite{zhu2009p}. One important property of SVMs is that the resulting classifier only uses a few training points, which are called \textit{support vectors}, to classify a new data point. \\
In addition to performing linear classification, SVMs can perform a non-linear classification which finds a hyperplane that is a non-linear function of the input variable. To do so, we implicitly map an input variable into high-dimensional feature spaces, a process which is called \textit {kernel trick} \cite{cristianini2000i}. In addition to performing binary classification, SVMs can perform multiclass classification. There are various ways to do so, such as One-vs-all (OVA) SVM, All-vs-all (AVA) SVM \cite{barber2012}, Structured SVM \cite{yu2009l}, and the Weston and Watkins \cite{weston1999} version.\\
SVMs are among the best \textit{off-the-shelf}, supervised learning models that are capable of effectively dealing with high-dimensional data sets and are efficient regarding memory usage due to the employment of support vectors for prediction. One significant drawback of this model is that it does not directly provide probability estimates. When given a solved SVM model, its parameters are difficult to interpret \cite{hsu2002c}. SVMs are of use in many real-world applications such as hand-written character recognition \cite{kim2003c}, image classification \cite{foody2004r}, and protein classification\cite{leslie2002s}. Finally, we should note that SVMs can be trained in an online fashion, which is addressed in \cite{poggio2001}. 
\cite{cite42} proposed a method on the Intel Lab Dataset. This data set consist of four environmental variables (Temperature, Voltage, Humidity, light) collected through S4 Mica2Dot sensors over 36 days at per-second rate.

\subsection{Regression}

\subsubsection{Linear Regression}
\label{LinReg}
In linear regression the objective is to learn a function $f(x, w)$. This is a mapping $f : \phi(x) \rightarrow y$ and is a linear combination of a fixed set of a linear or nonlinear function of the input variable denoted as $\phi_i(x)$, called a \textit{basis function}. The form of $f(x, w)$ is
\begin{equation}
\label{eqn:01}
f(x, w) =  \phi(x)^T w
\end{equation}
where $w$ is the weight vector or matrix $w=(w_1,\ldots,w_D)^T$, and  $\phi=(\phi_1,\ldots,\phi_D)^T$. There exists a broad class of basis functions such as \textit{polynomial}, \textit{gaussian radial}, and \textit{sigmoidal} basis functions which should be chosen with respect to the application \cite{montgomery2015,neter1996}.\\

For training the model, there exists a range of approaches: Ordinary Least Square, Regularized Least Squares, Least-Mean-Squares (LMS) and Bayesian Linear Regression. Among them, LMS is of particular interest since it is fast, scaleable to large data sets and learns the parameters online by applying the technique of \textit{stochastic gradient descent}, also known as \textit{sequential gradient descent} \cite{seber2012,bishop1}.\\

By using proper basis functions, it can be shown that arbitrary nonlinearities in the mapping from the input variable to output variable can be modeled. However, the assumption of fixed basis functions leads to significant shortcomings with this approach. For example, the increase in the dimension of the input space is coupled with  rapid growth in the number of basis functions \cite{bishop1,neter1996,murphy2012}. 
Linear regression can process at a high rate; \cite{cite45} use this algorithm to analyze and predict the energy usage of buildings.

\subsubsection{Support Vector Regression}
\label{RVM}
The SVM model described in Section \ref{SVM} can be extended to solve regression problems through a process called Support Vector Regression (SVR). Analogous to support vectors in SVMs, the resulting SVR model depends only on a subset of the training points due to the rejection of training points that are close to the model prediction \cite{smola1997s}. Various implementations of SVR exist such as epsilon-support vector regression and nu-support vector regression \cite{smola2004t}.
Authors in \cite{cite41} proposed a hybrid method to have accurate temperature and humidity data prediction.

\subsection{Combining Models}
\subsubsection{Classification and Regression Trees}
\label{Tree}
In classification and regression trees (CART), the input space is partitioned into axis-aligned cuboid regions $R_k$, and then a separate classification or regression model is assigned to each region in order to predict a label for the data points which fall into that region \cite{breiman1984}. Given a new, unseen input vector (data point) $x$, the process of predicting the corresponding target label can be explained by traversal of a binary tree corresponding to a sequential decision-making process. An example of a model for classification is one that predicts a particular class over each region and for regression, a model is one that predicts a constant over each region. To formulate the classification task, we denote a class variable by a discrete random variable $t$ and the predicted class label for $x$ by $y$. The  classification task takes the form of
\begin{equation}
\label{eqn:Tree1}
\begin{split}
&p (t = c | k) = \frac{1}{|R_k|} \sum_{i \in R_k} \mathbbm{1}{(t_i = c)}, 
\\ & y = \argmax_cp(t = c | x) = \argmax_cp(t = c | k)
\end{split}
\end{equation}
where $\mathbbm{1}{(.)}$ is the indicator function described in Section \ref{KNN}. This equation means $x$ will be labeled by the most common (mode) label in it's corresponding region \cite{loh2011}.\\

To formulate the regression task, we denote the value of the output vector by $t$ and the predicted output vector for $x$ by $y$. The  regression task is expressed as
\begin{equation}
\label{eqn:Tree2}
y = \frac{1}{|R_k|} \sum_{i \in R_k} t_i
\end{equation}
i.e., the output vector for $x$ will be the mean of the output vector of data points in it's corresponding region \cite{loh2011}.\\

\begin{algorithm}[t]
\label{alg2}
\SetAlgoNoLine
\KwIn{labeled training data set $D = \{(x_i,y_i)\}_{i=1}^N$.}
\KwOut{Classification or regression tree.}
\Call{fitTree}{$0$, $D$, $node$}\\

\begin{algorithmic}
  \Function{fitTree}{$depth$, $R$, $node$}
  \State {
  \eIf{the task is classification}{
  	$node$.prediction $:=$ most common label in $R$ 
  }{
  	$node$.prediction $:=$ mean of the output vector of the data points in $R$
  }
  $(i^*,z^*, R_L, R_R) :=$ \Call{split}{$R$}
  
  \If{worth splitting and stopping criteria is not met}{
  	$node$.test $:=$ $x_{i^*} < z^*$
    
    $node$.left $:=$ \Call{fitTree}{$depth + 1$, $R_L$, $node$}
    
    $node$.right $:=$ \Call{fitTree}{$depth + 1$, $R_R$, $node$}
  }
  \Return node
}
\EndFunction
\end{algorithmic}
\caption{Algorithm for Training CART}
\end{algorithm}

To train CART, the structure of the tree should be determined based on the training set. This means determining the split criterion at each node and their threshold parameter value. Finding the optimal tree structure is an NP-complete problem, therefore a \textit{greedy heuristic} which grows the tree top-down and chooses the best split node-by-node is used to train CART. To achieve better generalization and reduce overfilling some stopping criteria should be used for growing the tree. Possible stopping criterion are: the maximum depth reached,  whether the distribution in the branch is pure, whether the benefit of splitting is below a certain threshold, and whether the number of samples in each branch is below the criteria threshold. Moreover, after growing the tree, a pruning procedure can be used in order to reduce overfitting, \cite{prasad2006,bishop1,murphy2012}. Algorithm \ref{alg2} describes how to train  CART.\\

The major strength of CART is it's human interpretability due to its tree structure. Additionally, it is fast and scalable to large data sets; however, it is very sensitive to the choice of the training set \cite{trevor2001}. Another shortcoming with this model is unsmooth labeling of the input space since each region of input space is associated with exactly one label \cite{loh2011,bishop1}.
\cite{cite55} proposes an efficient and effective data-mining procedure that models the travel patterns of transit riders in Beijing, China.

\subsubsection{Random Forests}
\label{RandF}
In random forests, instead of training a single tree, an army of trees are trained. Each tree is trained on a subset of the training set, chosen randomly along with replacement, using a randomly chosen subset of $M$ input variables (features) \cite{breiman2001}. From here, there are two scenarios for the predicted label of a new, unseen data point: (1) in classification tasks; it is used as the mode of the labels predicted by each tree; (2)  in regression tasks it is used as the mean of the labels predicted by each tree. There is a tradeoff between different values of $M$. A value of $M$ that is too small leads to random trees with penniless prediction power, whereas a value of  $M$ that is too large leads to very similar random trees.\\

Random forests have very good accuracy but at the cost of losing human interpretability \cite{caruana2006}. Additionally, they are fast and scalable to large data sets and have many real-world applications such as body pose recognition \cite{Shotton2013} and body part classification.

\subsubsection{Bagging}
\label{Bagg} 
\textit{Bootstrap aggregating}, also called bagging, is an ensemble technique that aims to improve the accuracy and stability of machine learning algorithms and reduce overfitting. In this technique, $K$ new $M$ sized training sets are generated by randomly choosing data points from the original training set with replacement. Then, on each new generated training set, a machine learning model is trained, and the predicted label of a new, unseen data point is the mode of the predicted labels by each model in the case of classification tasks and is  the mean in the case of regression tasks. There are various machine learning models such as CART and neural networks, for which the bagging technique can improve the results. However, bagging degrades the performance of stable models such as KNN
 \cite{breiman1996b}. Examples of  practical applications include customer attrition prediction \cite{au2010m} and preimage learning \cite{sahu2011i,shinde2014}.


\subsection{Clustering}
\label{Cluster}
\subsubsection{K-means}
\label{KMeans}
In K-means algorithm, the objective is to cluster the unlabeled data set into a given $K$ number of clusters (groups) and data points belonging to the same cluster must have some similarities. In the classical K-means algorithm, the distance between data points is the measure of similarity. Therefore, K-means seeks to find a set of $K$ cluster centers, denoted as $\{s_1,\ldots,s_k\}$, which minimize the distance between data points and the nearest center \cite{Ng1}. In order to denote the assignment of data points to the cluster centers, we use a set of binary indicator variables $\pi_{nk}\in\{0, 1\}$; so that if data point $x_n$ is assigned to the cluster center $s_k$, then $\pi_{nk}=1$. We formulate the problem as follows:
\begin{equation}
\label{eqn:02}
\begin{aligned}
& \underset{s, \pi}{\text{minimize}}
& & \sum_{n=1}^{N}\sum_{k=1}^{K} \pi_{nk} \|x_n-s_k\|^2 \\
& \text{subject to}
& & \sum_{k=1}^{K} \pi_{nk} = 1, \; n = 1,\ldots, N.
\end{aligned}
\end{equation}
Algorithm \ref{alg1} describes how to learn the optimal cluster centers $\{s_k\}$ and the assignment of the data points $\{\pi_{nk}$\}.\\
\begin{algorithm}[t]
\label{alg1}
\SetAlgoNoLine
\KwIn{$K$, and unlabeled data set $\{x_1,\ldots,x_N\}$.}
\KwOut{Cluster centers $\{s_k\}$ and the assignment of the data points $\{\pi_{nk}\}$.}
Randomly initialize $\{s_k\}$.

\Repeat{$\{\pi_{nk}\}$ or $\{s_k\}$ don’t change}{
	\For{$n:=1$ to $N$}{
    	\For{$k:=1$ to $K$}{
        	\eIf{$k = \argmin_i\|s_i - x_i\|^2$}{
            	$\pi_{nk} := 1 $
            }
            { 
            	$\pi_{nk} := 0 $
            }
    	}
    }
	\For{$k:=1$ to $K$}{
    	$s_k := \frac{\sum_{n=1}^{N}x_n\pi_{nk}}{\sum_{n=1}^{N}\pi_{nk}}$
    }
}
\caption{K-means Algorithm}
\end{algorithm}

In practice, K-means is a very fast and highly scalable algorithm. Moreover, there is an  stochastic, online version of K-means \cite{jumutc2015}. However, this approach has many limitations due to the use of Euclidean distance as the measure of similarity. For instance, it has limitations on the types of data variables that can be considered and cluster centers are not robust against outliers. Additionally, the K-means algorithm assigns each data point to one, and only one of the clusters which may lead to inappropriate clusters in some cases \cite{likas2003}.  
\cite{cite48} use MapReduce to analyze the numerous small data sets and proposes a cluster strategy for high volume of small data based on the k-means algorithm.
\cite{cite55} applied K-Means++ to cluster and classify travel pattern regularities. \cite{cite59} introduced real-time event processing and clustering algorithm for analyzing sensor data by using the OpenIoT1 middleware as an interface for innovative analytical IoT services.


\subsubsection{Density-Based Spatial Clustering of Applications with Noise}
\label{DBSCAN}
In a density-based spatial clustering of applications with noise (DBSCAN) approach, the objective is to cluster a given unlabeled data set based on the density of its data points. In this model, groups of dense data points (data points with many close neighbors) are considered as clusters and data points in regions with low-density are considered as outliers \cite{kriegel2011d}. \cite{ester1996d} present an algorithm to train a DBSCAN model.\\    

In practice, DBSCAN is efficient on large datasets and is fast and robust against outliers. Also, it is capable of detecting clusters with an arbitrary shape  (i.e., spherical, elongated, and linear). Moreover, the model determines the number of clusters based on the density of the data points, unlike K-means which requires the number of clusters to be specified \cite{ester1996d}. However, there are some disadvantages associated with DBSCAN. For example, in the case of a data set with large differences in densities, the resulting clusters are destitute. Additionally, the performance of the model is very sensitive to the distance metric that is used for determining if a region is dense \cite{campello2013d}. It is worth, however, noting that DBSCAN is among the most widely used clustering algorithms with numerous real world applications such as anomaly detection in temperature data \cite{ccelik2011an} and X-ray crystallography \cite{ester1996d}.
Authors in \cite{cite42} believe that knowledge discovery in data streams is a valuable task for research, business, and community. 
They applied Density-based clustering algorithm DBSCAN on a data stream to reveal the number of existing classes and subsequently label of the data. Also In \cite{cite47} this algorithm used  to find the arbitrary shape of the cluster. DBSCAN algorithm produces sets of clusters with arbitrary shape and outliers objects.

\subsection{Feature Extraction}
\label{featureE}
\subsubsection{Principal Component Analysis}
\label{PCA}
In principle component analysis (PCA), the objective is to orthogonally project data points onto an $L$ dimensional linear subspace, called the \textit{principal subspace}, which has the maximal projected variance \cite{hotelling1933,abdi2010}. Equivalently, the objective can be defined as finding a complete orthonormal set of $L$ linear basis M-dimensional vectors $\{w_j\}$ and the corresponding linear projections of data points $\{z_{nj}\}$ such that the average reconstruction error\\
\begin{equation}
\label{eqn:05}
\begin{split}
&J = \frac{1}{N}\sum_n\|\tilde{x}_n - x_n\|^2, 
\\& \tilde{x}_n = \sum_{j=1}^Lz_{nj}w_j + \bar{x}
\end{split}
\end{equation}
is minimized, where $\bar{x}$ is the average of all data points \cite{pearson1901,bishop1}.\\

\begin{algorithm}[t]
\label{algPCA}
\SetAlgoNoLine
\KwIn{$L$, and input vectors of an unlabeled or labeled data set $\{x_1,\ldots,x_N\}$.}
\KwOut{The projected data set $\{z_1,\ldots,z_N\}$, and basis vectors $\{w_j\}$ which form the principal subspace.}

$\bar{x} := \frac{1}{N} \sum_nx_n$

$S := \frac{1}{N}\sum_n(x_n-\bar{x})(x_n-\bar{x})^T$

$\{w_j\} := $ the $L$ eigenvectors of $S$ corresponding to the $L$ largest eigenvalues.

\For{$n:=1$ to $N$}{
	\For{$j:=1$ to $L$}{
    	$z_{nj} := (x_n-\bar{x})^Tw_j$
    }
}

\caption{PCA Algorithm}
\end{algorithm}

Algorithm \ref{algPCA} describes how the PCA technique achieves these objectives. Depending on how  $\{w_1,\ldots,w_L\}$ is calculated, the PCA algorithm can have different run times i.e., $O(M^3)$, $O(LM^2)$, $O(NM^2)$ and $O(N^3)$ \cite{golub2012matrix,bishop1,sirovich1987turbulence}. In order to deal with high dimensional data sets, there is a different version of the PCA algorithm which is based on the iterative \textit{Expectation Maximization} technique. In this algorithm, the covariance matrix of the dataset is not explicitly calculated, and its most computationally demanding steps are $O(NML)$. In addition, this algorithm can be implemented in an online fashion, which can also be advantageous in cases where  $M$ and $N$ are large \cite{jolliffe2002,murphy2012}.\\

PCA is one of the most important preprocessing techniques in machine learning. Its application involves data compression, whitening, and data visualization. Examples of its practical applications are face recognition, interest rate derivatives portfolios, and neuroscience. Furthermore, there exists a kernelized version of PCA, called KPCA which can find nonlinear principal components \cite{bro2014,jolliffe2002}. 

\subsubsection{Canonical Correlation Analysis}
\label{CCA}
Canonical correlation analysis (CCA), is a linear dimensionality reduction technique which is closely related to PCA. Unlike PCA which deals with one variable, CCA deals with two or more variables and its objective is to find a corresponding pair of highly cross-correlated linear subspaces so that within one of the subspaces there is a correlation between each component and a single component from the other subspace. The optimal solution can be obtained by solving a generalized eigenvector problem \cite{hotelling1936,bach2002,bishop1}.
\cite{cite51} compared PCA and CCA for detecting intermittent faults and masking failures of the indoor environments. 

\subsection{Neural Network}
One of the shortcomings of linear regression is that it requires deciding the types of basis functions. It is often hard to decide the optimal basis functions. Therefore, in neural networks we fix the number of basis functions but we let the model learn the parameters of the basis functions. There exist many different types of neural networks with different architectures, use cases, and applications. In subsequent subsections, we discuss the successful models used in smart data analysis. Note that, neural networks are fast to process new data since they are compact models; on the contrary, however, they usually need the high amount of computation in order to be trained. Moreover, they are easily adaptable to regression and classification problems \cite{eberhart2014,glorot2010understanding}.

\subsubsection{Feed Forward Neural Network}
\label{FFNN}
\begin{figure}
\centerline{\includegraphics[width=0.5\linewidth]{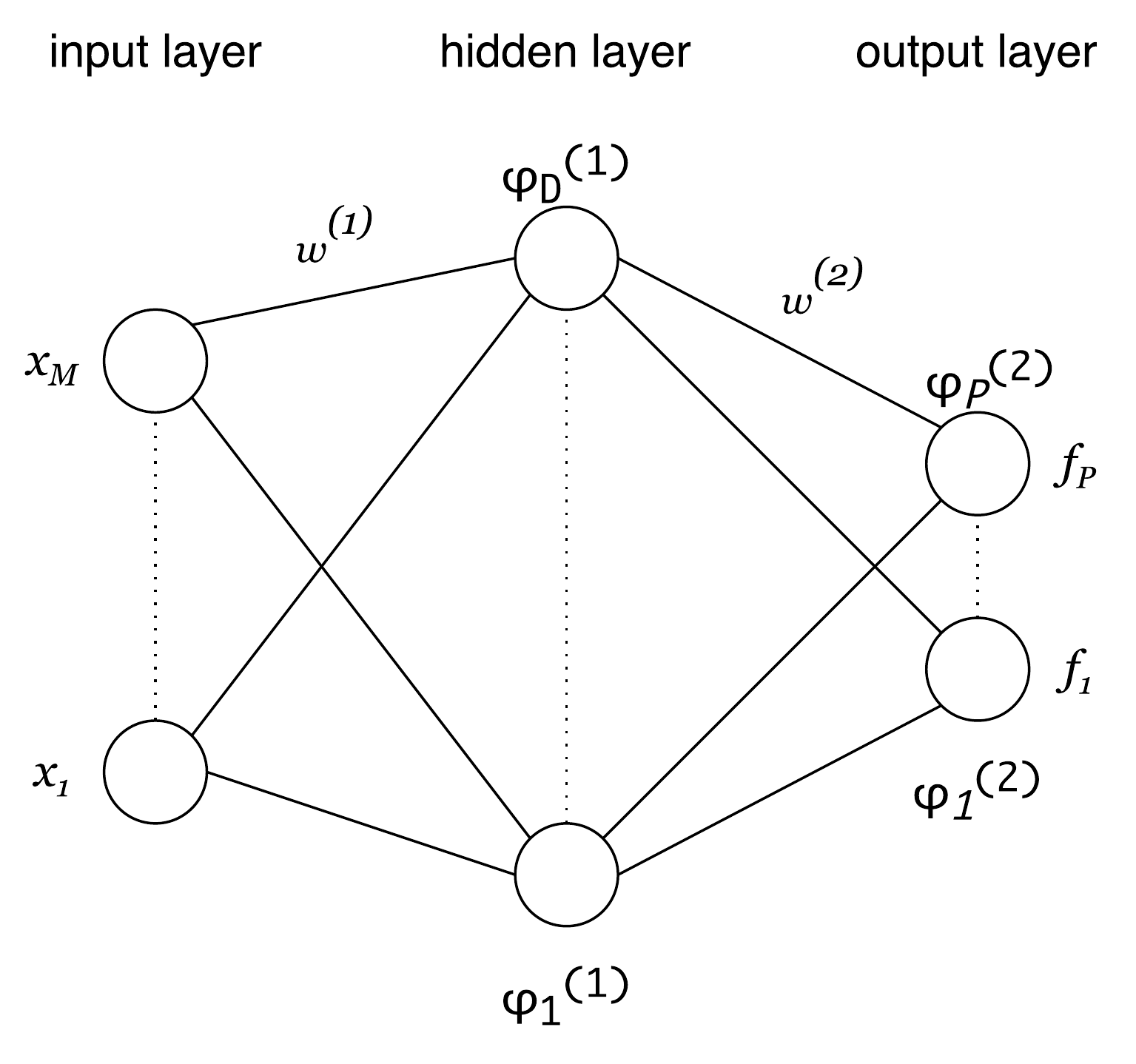}}

\caption{A two layers feed forward neural network. Note that each output neuron is connected to each input neuron, i.e., it is a fully connected neural network.}
\label{fig:one}
\end{figure}

Feed Forward Neural Networks (FFNN), also known as \textit{multilayer perceptrons} (MLP), are the most common type of neural networks in practical applications. To explain this model we begin with a simple two layer FFNN model. Assume that we have $D$ basis functions and our objective is to learn the parameters of these basis functions together with the function $f$ discussed in Section \ref{LinReg}. The form of the classification or regression task is

\begin{equation}
\label{eqn:04}
f(x, w^{(1)}, w^{(2)}) =  \phi^{(2)}(\phi^{(1)}(x^T w^{(1)})^T w^{(2)})
\end{equation}
where $w^{(1)}=(w^{(1)}_1,\ldots,w^{(1)}_M)^T$, $\phi^{(1)} =(\phi^{(1)}_1,\ldots,\phi^{(1)}_D)^T$, $w^{(2)}=(w^{(2)}_1,\ldots,w^{(2)}_D)^T$, and $\phi^{(2)} =(\phi^{(2)}_1,\ldots,\phi^{(2)}_P)^T$. Figure \ref{fig:one} visualizes this FFNN model. The elements of input vector $x$ are units (neurons) in the input layer, $\phi^{(1)}_i$ are the units in the hidden layer, and $\phi^{(2)}_i$ are the units in the output layer which outputs $f$. Note that the activities of the units in each layer are a nonlinear function of the activities in the previous layer. In machine learning literature, $\phi(.)$ is also called \textit{activation function}. The activation function in the last layer is chosen with respect to the data processing task. For example, for regression task we use linear activation and for multiclass classification we use \textit{softmax} activation function \cite{Goodfellow-et-al-2016,eberhart2014,bishop1}.\\

With enough hidden units, an FFNN with at least two layers can approximate an arbitrary mapping from a finite input space to a finite output space \cite{cybenko1989,hornik1991,fukushima1980}. However, for an FFNN, finding the optimum set of weights $w$ is an NP-complete problem \cite{blum1992teaching}. To train the model, there is a variety range of learning methods such as stochastic gradient descent, adaptive delta, adaptive gradient, adaptive moment estimation, Nesterov's accelerated gradient and RMSprob. To improve the generalization of the model and reduce overfitting, there are a range of methods such as weight decay, weight-sharing, early stopping, Bayesian fitting of neural nets, dropout, and generative pre-training \cite{he2015,lecun1998}.\\ 

A two layer FFNN has the properties of restricted representation and generalization. Moreover, compactly represented functions with $l$ layers may require exponential size with $l-1$ layers. Therefore, an alternative approach would be an FFNN with more than one hidden layers, i.e., a \textit{deep neural network}, in which different high-level features share low-level features \cite{lecun2015,glorot2010understanding}. Significant results with deep neural networks have led them to be the most commonly used classifiers in machine learning \cite{schmidhuber2015,Goodfellow-et-al-2016}.
\cite{cite57} present the method to forecast the states of IoT elements based on an artificial neural network. The presented architecture of the neural network is a combination of a multilayered perceptron and a probabilistic neural network. Also, \cite{cite13} use FFNN for processing the health data.

\subsection{Time Series and Sequential Data}
So far in this article, the discussed algorithms dealt with set of data points that are independent and identically distributed (i.i.d.). However, the set of data points are not i.i.d. for many cases, often resulting from time series measurements, such as the daily closing value of the Dow Jones Industrial Average and acoustic features at successive time frames. An example of non i.i.d set of data points in a context other than a time series is a character sequence in a German sentence. In these cases, data points  consist of sequences of $(x, y)$ pairs rather than being drawn i.i.d. from a joint distribution $p(\textbf{x},\textbf{y})$ and the sequences exhibit significant sequential correlation \cite{dietterich2002m,bishop1}.\\

In a \textit{sequential supervised learning} problem, when data points are sequential, we are given a training set $\{(x_i,y_i)\}^N_{i=1}$ consisting of $N$ samples and each of them is a pair of sequences. In each sample, $x_i = \langle x_{i,1},x_{i,2},\ldots,x_{i,T_i}\rangle$ and $y_i = \langle y_{i,1},y_{i,2},\ldots,y_{i,T_i}\rangle$. Given a new, unseen input sequence $x$, the goal is to predict the desired output sequence $y$. Moreover, there is  a closely related problem, called a \textit{time-series prediction} problem, in which the goal is to predict the desired $t+1$\textsuperscript{st} element of a sequence $\langle y_1,\ldots, y_t\rangle$. The key difference between them is that unlike sequential supervised learning, where the entire sequence $\langle x_1,\ldots,x_T\rangle$ is available prior to any prediction, in time-series prediction only the prefix of the sequence, up to the current time $t+1$, is available. In addition, in sequential supervised learning, the entire output sequence $y$ has to be predicted, whereas in time-series prediction, the true observed values of the output sequence up to time $t$ are given. It is worth noting that there is another  closely-related task, called \textit{sequence classification}, in which the goal is to predict the desired, single categorical output $y$ given an input sequence $x$ \cite{dietterich2002m}.\\

There are a variety of machine learning models and methods which can deal with these tasks. Examples of these models and methods are hidden Markov models \cite{baum1967i,rabiner1989t}, sliding-window methods \cite{sejnowski1987p}, Kalman filter \cite{kalman1961n}, conditional random fields \cite{lafferty2001c}, recurrent neural networks \cite{williams1989l,Goodfellow-et-al-2016}, graph transformer networks \cite{lecun1998}, and maximum entropy Markov models \cite{mccallum2000m}. In addition, sequential time series and sequential data exists in many real world applications, including  speech recognition \cite{sak2014l}, handwriting recognition \cite{graves2009n}, musical score following \cite{pardo2005m}, and information extraction \cite{mccallum2000m}.


\subsection{Anomaly Detection}
The problem of identifying items or patterns in the data set that do not conform to other items or an expected pattern is referred to as \textit{anomaly detection} and these unexpected patterns are called anomalies, outliers, novelties, exceptions, noise, surprises, or deviations \cite{hodge2004s,chandola2009a}.\\

There are many challenges in the task of  anomaly detection  which distinguish it from a binary classification task. For example, an anomalous class is often severely underrepresented in the training set. In addition, anomalies are much more diverse than the behavior of the normal system and are sparse by  nature \cite{milacski2015r,chandola2009a}.\\

There are three broad categories of anomaly detection techniques based on the extent to which the labels are available. In \textit{supervised anomaly detection} techniques, a binary (abnormal and normal) labeled data set is given, then,  a binary classifier is trained; this should deal with the problem of the \textit{unbalanced data set} due to the existence of few data points with the abnormal label. \textit{Semi-supervised anomaly detection} techniques require a training set that contains only normal data points. Anomalies are then detected by building the normal behavior model of the system and then testing the likelihood of the generation of the test data point by the learned model. \textit{Unsupervised anomaly detection} techniques deal with an unlabeled data set by making the implicit assumption that the majority of the data points are normal \cite{chandola2009a}.\\

Anomaly detection is of use in many real world applications such as system health monitoring, credit card fraud detection,  intrusion detection \cite{denning1987i}, detecting eco-system disturbances, and military surveillance. Moreover, anomaly detection can be used as a preprocessing algorithm for removing outliers from the data set, that  can significantly improve the performance of the subsequent machine learning algorithms, especially in  supervised learning tasks \cite{tomek1976e,smith2011i}. In the following subsection we shall explain \textit{one-class support vector machines}  one of the most popular techniques for anomaly detection.
\cite{cite58} build a novel outlier detection algorithm that uses statistical techniques to identify outliers and anomalies in power datasets collected from smart environments.

\subsubsection{One-class Support Vector Machines}
\label{OCSVM}
One-class support vector machines (OCSVMs) are a semi-supervised anomaly detection technique and are an extension of the SVMs discussed in Section \ref{SVM} for unlabeled data sets. Given a training set drawn from an underlying probability distribution $P$, OCSVMs aim to estimate a subset $S$ of the input space such that the probability that a drawn sample from $P$ lies outside of $S$ is bounded by a fixed value between $0$ and $1$. This problem is approached by learning a binary function $f$ which captures the input regions where the probability density lives. Therefore, $f$ is negative in the complement of $S$. The functional form of $f$ can be computed by solving a quadratic programming problem \cite{scholkopf2001e,ratsch2002c}.\\

One-class SVMs are  useful in many anomaly detection applications, such as anomaly detection in sensor networks \cite{rajasegarar2010c}, system called intrusion detection \cite{heller2003o}, network intrusions detection \cite{zhang2015an}, and anomaly detection in wireless sensor networks \cite{zhang2009ad}.
\cite{cite47} reviewed different techniques of stream data outlier detection and their issues in detail. \cite{cite52} use One-class SVM  to detect anomalies by modeling the complex normal patterns in the data.

In the following section, we discussed how to overcome the challenges of applying machine learning algorithms to the IoT smart data.

\section{Discussion on taxonomy of machine  learning  algorithms}
\label{Disc}
In order to draw the right decisions for smart data analysis, it is necessary to determine which one of the tasks whether structure discovery, finding unusual data points, predicting values, predicting categories, or feature extraction should be accomplished.

To discover the structure of data, the one that faces with the unlabeled data, the clustering algorithms can be the most appropriate tools. K-means described in \ref{KMeans} is the well-known and frequently applied clustering algorithm, which can handle a large volume of data with a broad range of data types. \cite{cite46,cite47} proposed a method for applying K-means algorithm in managing the Smart City and Smart Home data.
DB-scan described in \ref{DBSCAN} is another clustering algorithm to discover the structure of data from the unlabeled data which is applied in \cite{cite42,cite47,cite55} to cluster Smart Citizen behaviors.

To find unusual data points and anomalies in smart data, two important algorithms are applied. One class Support Vector Machine and PCA based anomaly detection methods explained in \ref{PCA} which have the ability to train anomaly and noisy data with a high performance. \cite{cite47,cite52} applied the One class SVM monitor and find the human activity anomalies.

In order to predict values and classification of sequenced data, Linear regression and SVR described in \ref{LinReg} and \ref{RVM} are the two frequently applied algorithms. The objective of the models applied in these algorithms is to process and train data of high velocity. For example \cite{cite45,cite41} applied linear regression algorithm for real-time prediction. Another fast training algorithm is the classification and regression tree described in \ref{Tree}, applied in classifying Smart Citizen behaviors \cite{cite45,cite55}.

To predict the categories of the data, neural networks are proper learning models for function approximation problems. Moreover, because the smart data should be accurate and it takes a long time to be trained, the multi-class neural network can be an appropriate solution. For instance, Feed Froward Neural Network explained in \ref{FFNN} applied to reduce energy consumption in future by predicting how the data in future will be generated and how the redundancy of the data would be removed \cite{cite13,cite57,cite60}. SVM explained in \ref{SVM} is another popular classification algorithm capable of handling massive amounts of data and classify their different types. Because SVM solves the high volume and the variety types of data, it is commonly applied in most smart data processing algorithms. For example, \cite{cite42,cite45} applied SVM to classify the traffic data.

PCA and CCA described in \ref{PCA} and \ref{CCA} are the two algorithms vastly applied in extracting features of the data. Moreover, CCA shows the correlation between the two categories of the data. A type of PCA and CCA are applied to finding the anomalies.
\cite{cite51} applied PCA and CCA to monitor the public places and detect the events in the social areas.


The chosen algorithm should be implemented and developed to make right decisions.

A sample implemented code is available from the open source GitHub license at https://github.com/mhrezvan/SVM-on-Smart-Traffic-Data


\section{Research trends and open issues}
\label{Research-trends}
As discussed before, data analysis have a significant contribution to IoT; therefore to applied a full potential of analysis to extract new insights from data, IoT must overcome some major problems. These problems can be categorized in three different types.
\subsection{IoT Data Characteristics} 
Because the data are the basis of extracting knowledge, it is vital to have high quality information. This condition can affect the accuracy of knowledge extraction in a direct manner. Since IoT produces high volume, fast velocity, and varieties of data, preserving the data quality is a hard and challenging task. Although many solutions have been and are being introduced to solve these problems, none of them can handle all aspects of data characteristics in an accurate manner because of the distributed nature of Big Data management solutions and real-time processing platforms. 
The abstraction of IoT data is low, that is, the data that comes from different resources in IoT are mostly of raw data and not sufficient enough for analysis. A wide variety of solutions are proposed, while most of them need further improvements. For instance, semantic technologies tend to enhance the abstraction of IoT data through annotation algorithms, while they need more efforts to overcome its velocity and volume.
\subsection{IoT Applications}
IoT applications have different categories according to their unique attributions and features. Certain issues should be proposed in running data analysis in IoT applications in an accurate manner. First, the privacy of the collected data is very critical, since data collection process can include personal or critical business data, which is inevitable to solve the privacy issues. Second, according to the vast number of resources and simple-designed hardware in IoT, it is vital to consider security parameters like network security, data encryption, etc. Otherwise, by ignoring the security in design and implementation, an infected network of IoT devices can cause a crisis.
\subsection{IoT Data Analytics Algorithms}
According to the smart data characteristics, analytic algorithms should be able to handle  Big Data, that is, IoT needs  algorithms that can analyze the data which comes from a variety of sources in real time. Many attempts are made to address this issue. For example, deep learning algorithms, evolutionized form of neural networks can reach to a high accuracy rate if they have enough data and time. Deep learning algorithms can be easily influenced by the smart noisy data, furthermore, neural network based algorithms lack interpretation, this is, data scientists can not understand the reasons for the model results. In the same manner, semi-supervised algorithms which model the small amount of labeled data with a large amount of unlabeled data can assist IoT data analytics as well.

\begin{table}[htbp]
\footnotesize
   \centering
  
   \caption{Overview of Applying Machine Learning Algorithm to the Internet of Things Use Cases}
   \label{tab:three}
\begin{tabular}{|C{3.1cm}|C{2.6cm}|C{6cm}|C{1.8cm}|}
\hline

\textbf{\specialcell{Machine learning \\Algorithm} } & \textbf{\specialcell{IoT, Smart City\\use cases}}& \textbf{\specialcell{Metric to Optimize}} & 
\textbf{References}
\\\hline

\specialcell {Classification }&\specialcell{Smart Traffic  } & \specialcell{Traffic Prediction,\\Increase Data Abbreviation} & \specialcell {\cite{cite43}\\ \cite{cite4}\\ } \\\hline

\specialcell {Clustering} &\specialcell{Smart Traffic,\\ Smart Health } & \specialcell{Traffic Prediction,\\ Increase Data Abbreviation}&\specialcell {\cite{cite43} \cite{cite4}\\ \cite{cite49}} \\\hline

\specialcell {Anomaly Detection} & \specialcell{Smart Traffic,\\Smart Environment } & \specialcell{Traffic Prediction, Increase Data Abbreviation, \\Finding Anomalies in Power Dataset }& \specialcell {\cite{cite43} \cite{cite4}\\ \cite{cite58} }\\\hline

\specialcell {Support Vector \\Regression} & \specialcell{Smart Weather \\ Prediction } & Forecasting&\cite{cite41}\\\hline

\specialcell {Linear Regression} & \specialcell{Economics,\\ Market analysis,\\ Energy usage }  &\specialcell{Real Time Prediction, \\Reducing Amount of Data }& \specialcell{\cite{cite45}  \\ \cite{cite60}}\\\hline

\specialcell {Classification \\ and Regression Trees} &Smart Citizens & \specialcell{Real Time Prediction, \\ Passengers Travel Pattern} & \specialcell{\cite{cite45} \\ \cite{cite55}}\\\hline


\specialcell {Support Vector Machine} & \specialcell{All Use Cases } &\specialcell{Classify Data,\\Real Time Prediction } & \specialcell{ \cite{cite42} \\ \cite{cite45}\\} \\\hline

\specialcell {K-Nearest Neighbors} &\specialcell {Smart Citizen \\   } & \specialcell{Passengers' Travel Pattern,\\ Efficiency of the Learned Metric}  & \specialcell {\cite{cite55} \\ \cite{cite56}} \\\hline

\specialcell {Naive Bayes} & \specialcell{Smart Agriculture,\\ Smart Citizen} & \specialcell{Food Safety, Passengers Travel Pattern,\\ Estimate the Numbers of Nodes } & \specialcell{\cite{cite50}  \cite{cite55}  \\ \cite{cite60}}\\\hline



\specialcell{K-Means}  & \specialcell{Smart City,\\ Smart Home,\\ Smart Citizen, \\ Controlling Air \\ and Traffic} & \specialcell {Outlier Detection, fraud detection, \\ Analyze Small Data set,\\ Forecasting Energy Consumption, \\ Passengers Travel Pattern, Stream Data Analyze}&  \specialcell{\cite{cite46}  \cite{cite47}  \\ \cite{cite48} \cite{cite40} \\ \cite{cite55} \cite{cite59}}  \\\hline

\specialcell{Density-Based  Clustering }  & Smart Citizen  & \specialcell {Labeling Data, Fraud Detection,\\ Passengers Travel Pattern} & \specialcell{\cite{cite42}  \cite{cite47} \\ \cite{cite55}} \\\hline


\specialcell {Feed Forward \\ Neural Network }&Smart Health & \specialcell {Reducing  Energy Consumption, Forecast the \\ States of Elements, \\ Overcome the  Redundant Data  and Information}&\specialcell{ \cite{cite13}  \cite{cite57} \\ \cite{cite60}}\\\hline

\specialcell {Principal \\ Component Analysis} &  \specialcell {Monitoring\\ Public Places}  & Fault Detection  &  \cite{cite51}\\\hline

\specialcell {Canonical \\ Correlation Analysis} & \specialcell {Monitoring\\ Public Places} & Fault Detection & \cite{cite51} \\\hline

\specialcell {One-class Support} Vector Machines &\specialcell{Smart Human \\ Activity Control} & \specialcell {Fraud Detection,  Emerging Anomalies in the data } &\specialcell{ \cite{cite47}   \cite{cite52}}\\\hline

\end{tabular}
\end{table}%

\section{Conclusions}

IoT consists of a vast number of devices with  varieties that are connected to each other and transmit huge amounts of data. The Smart City is one of the most important applications of IoT and provides different services in domains like energy, mobility, and urban planning. These services can be enhanced and optimized by analyzing the smart data collected from these areas. In order to extract knowledge from  collected data, many data analytic algorithms can be applied. Choosing a proper algorithm for specific IoT and Smart City application is an important issue. In this article, many IoT data analytic studies are reviewed to address this issue. Here three facts should be considered in applying data analytic algorithms to smart data. The first fact is that different applications in IoT and smart cities have their characteristics as the number of devices and types of the data that they generate; the second fact is that the generated data have specific features that should be realized. The third fact is that the taxonomy of the algorithms is another important point in applying data analysis to smart data. The findings in this article make the choice of proper algorithm for a particular problem easy.
The analytic algorithms are of eight categories, described in detail. This is followed by reviewing application specifics of Smart City use cases. The data characteristics and quality of smart data are described in detail. In the discussion section, how the data characteristics and application specifics can lead to choosing a proper data analytic algorithms is reviewed. In the future trend section the recent issues and the future path for research in the field of smart data analytics are discussed.





\section*{Acknowledgments}

The authors would like to thank Dr. Mohammadsaeid Ehsani and Mr. Ajit Joakar for their comments on the draft of the paper. We also thank Dr. Alireza Ahrabian and Utkarshani Jiamini for reviewing our paper.  
\section*{References}
\bibliographystyle{ACM-Reference-Format-Journals}
\bibliography{acmsmall-sample-bibfile}

\begin{thebibliography}{100}
\expandafter\ifx\csname url\endcsname\relax
  \def\url#1{\texttt{#1}}\fi
\expandafter\ifx\csname urlprefix\endcsname\relax\def\urlprefix{URL }\fi
\expandafter\ifx\csname href\endcsname\relax
  \def\href#1#2{#2} \def\path#1{#1}\fi

\bibitem{cite61}
L.~Atzori, A.~Iera, G.~Morabito, The internet of things: A survey, Computer
  networks 54~(15) (2010) 2787--2805.

\bibitem{cite3}
C.~Cecchinel, M.~Jimenez, S.~Mosser, M.~Riveill, An architecture to support the
  collection of big data in the internet of things, in: 2014 IEEE World
  Congress on Services, IEEE, 2014, pp. 442--449.

\bibitem{weiser1999c}
M.~Weiser, The computer for the 21st century., Mobile Computing and
  Communications Review 3~(3) (1999) 3--11.

\bibitem{sheth2010c}
A.~Sheth, Computing for human experience: Semantics-empowered sensors,
  services, and social computing on the ubiquitous web, IEEE Internet Computing
  14~(1) (2010) 88--91.

\bibitem{cite68}
J.~Manyika, M.~Chui, B.~Brown, J.~Bughin, R.~Dobbs, C.~Roxburgh, A.~H. Byers,
  Big data: The next frontier for innovation, competition, and productivity.

\bibitem{cite69}
A.~Sheth, Transforming big data into smart data: Deriving value via harnessing
  volume, variety, and velocity using semantic techniques and technologies, in:
  Data Engineering (ICDE), 2014 IEEE 30th International Conference on, IEEE,
  2014, pp. 2--2.

\bibitem{cite691}
A.~P. Sheth, Transforming big data into smart data for smart energy: Deriving
  value via harnessing volume, variety and velocity.

\bibitem{cite71}
A.~Sheth, Internet of things to smart iot through semantic, cognitive, and
  perceptual computing, IEEE Intelligent Systems 31~(2) (2016) 108--112.

\bibitem{cite29}
S.~Bin, L.~Yuan, W.~Xiaoyi, Research on data mining models for the internet of
  things, in: 2010 International Conference on Image Analysis and Signal
  Processing, IEEE, 2010, pp. 127--132.

\bibitem{cite30}
H.~Gonzalez, J.~Han, X.~Li, D.~Klabjan, Warehousing and analyzing massive rfid
  data sets, in: 22nd International Conference on Data Engineering (ICDE'06),
  IEEE, 2006, pp. 83--83.

\bibitem{cite31}
F.~Chen, P.~Deng, J.~Wan, D.~Zhang, A.~V. Vasilakos, X.~Rong, Data mining for
  the internet of things: literature review and challenges, International
  Journal of Distributed Sensor Networks 2015 (2015) 12.

\bibitem{cite32}
C.-W. Tsai, C.-F. Lai, M.-C. Chiang, L.~T. Yang, Data mining for internet of
  things: a survey, IEEE Communications Surveys \& Tutorials 16~(1) (2014)
  77--97.

\bibitem{cite70}
A.~Zanella, N.~Bui, A.~Castellani, L.~Vangelista, M.~Zorzi, Internet of things
  for smart cities, IEEE Internet of Things journal 1~(1) (2014) 22--32.

\bibitem{cite4}
Y.~Qin, Q.~Z. Sheng, N.~J. Falkner, S.~Dustdar, H.~Wang, A.~V. Vasilakos, When
  things matter: A survey on data-centric internet of things, Journal of
  Network and Computer Applications 64 (2016) 137--153.

\bibitem{cite5}
M.~Ma, P.~Wang, C.-H. Chu, Ltcep: Efficient long-term event processing for
  internet of things data streams, in: 2015 IEEE International Conference on
  Data Science and Data Intensive Systems, IEEE, 2015, pp. 548--555.

\bibitem{cite2}
P.~Barnaghi, A.~Sheth, The internet of things: The story so far, IEEE Internet
  of Things.

\bibitem{cite6}
Z.~Sheng, S.~Yang, Y.~Yu, A.~V. Vasilakos, J.~A. McCann, K.~K. Leung, A survey
  on the ietf protocol suite for the internet of things: Standards, challenges,
  and opportunities, IEEE Wireless Communications 20~(6) (2013) 91--98.

\bibitem{cite7}
F.~Bonomi, R.~Milito, J.~Zhu, S.~Addepalli, Fog computing and its role in the
  internet of things, in: Proceedings of the first edition of the MCC workshop
  on Mobile cloud computing, ACM, 2012, pp. 13--16.

\bibitem{cite11}
M.~Aazam, E.-N. Huh, Fog computing micro datacenter based dynamic resource
  estimation and pricing model for iot, in: 2015 IEEE 29th International
  Conference on Advanced Information Networking and Applications, IEEE, 2015,
  pp. 687--694.

\bibitem{cite12}
Y.~Shi, G.~Ding, H.~Wang, H.~E. Roman, S.~Lu, The fog computing service for
  healthcare, in: Future Information and Communication Technologies for
  Ubiquitous HealthCare (Ubi-HealthTech), 2015 2nd International Symposium on,
  IEEE, 2015, pp. 1--5.

\bibitem{cite13}
F.~Ramalho, A.~Neto, K.~Santos, N.~Agoulmine, et~al., Enhancing ehealth smart
  applications: A fog-enabled approach, in: 2015 17th International Conference
  on E-health Networking, Application \& Services (HealthCom), IEEE, 2015, pp.
  323--328.

\bibitem{cite14}
A.~Joakar, A methodology for solving problems with datascience for internet of
  things, DataScience for Internet of Things.

\bibitem{cite15}
A.~Papageorgiou, M.~Zahn, E.~Kovacs, Efficient auto-configuration of
  energy-related parameters in cloud-based iot platforms, in: Cloud Networking
  (CloudNet), 2014 IEEE 3rd International Conference on, IEEE, 2014, pp.
  236--241.

\bibitem{cite16}
L.~Wang, R.~Ranjan, Processing distributed internet of things data in clouds.,
  IEEE Cloud Computing 2~(1) (2015) 76--80.

\bibitem{cite17}
H.~Zhao, C.~Huang, A data processing algorithm in epc internet of things, in:
  Cyber-Enabled Distributed Computing and Knowledge Discovery (CyberC), 2014
  International Conference on, IEEE, 2014, pp. 128--131.

\bibitem{cite22}
R.~Petrolo, V.~Loscr{\`\i}, N.~Mitton, Towards a smart city based on cloud of
  things, a survey on the smart city vision and paradigms, Transactions on
  Emerging Telecommunications Technologies.

\bibitem{cite24}
E.~Von~Hippel, Democratizing innovation: The evolving phenomenon of user
  innovation, Journal f{\"u}r Betriebswirtschaft 55~(1) (2005) 63--78.

\bibitem{cite33}
D.~Puiu, P.~Barnaghi, R.~T{\"o}njes, D.~K{\"u}mper, M.~I. Ali, A.~Mileo, J.~X.
  Parreira, M.~Fischer, S.~Kolozali, N.~Farajidavar, et~al., Citypulse: Large
  scale data analytics framework for smart cities, IEEE Access 4 (2016)
  1086--1108.

\bibitem{cite23}
B.~Bowerman, J.~Braverman, J.~Taylor, H.~Todosow, U.~Von~Wimmersperg, The
  vision of a smart city, in: 2nd International Life Extension Technology
  Workshop, Paris, Vol.~28, 2000.

\bibitem{cite27}
J.~Pan, R.~Jain, S.~Paul, T.~Vu, A.~Saifullah, M.~Sha, An internet of things
  framework for smart energy in buildings: Designs, prototype, and experiments,
  IEEE Internet of Things Journal 2~(6) (2015) 527--537.

\bibitem{cite28}
J.~Torriti, Demand side management for the european supergrid: Occupancy
  variances of european single-person households, Energy Policy 44 (2012)
  199--206.

\bibitem{cite62}
Y.~Wang, J.~Yuan, X.~Chen, J.~Bao, Smart grid time series big data processing
  system, in: 2015 IEEE Advanced Information Technology, Electronic and
  Automation Control Conference (IAEAC), IEEE, 2015, pp. 393--400.

\bibitem{cite63}
S.~Karnouskos, T.~N. De~Holanda, Simulation of a smart grid city with software
  agents, in: Computer Modeling and Simulation, 2009. EMS'09. Third UKSim
  European Symposium on, IEEE, 2009, pp. 424--429.

\bibitem{cite64}
D.~R. Nagesh, J.~V. Krishna, S.~Tulasiram, A real-time architecture for smart
  energy management, in: Innovative Smart Grid Technologies (ISGT), 2010, IEEE,
  2010, pp. 1--4.

\bibitem{cite65}
T.~Robles, R.~Alcarria, D.~Mart{\'\i}n, A.~Morales, M.~Navarro, R.~Calero,
  S.~Iglesias, M.~L{\'o}pez, An internet of things-based model for smart water
  management, in: Advanced Information Networking and Applications Workshops
  (WAINA), 2014 28th International Conference on, IEEE, 2014, pp. 821--826.

\bibitem{cite66}
Z.~Zhao, W.~Ding, J.~Wang, Y.~Han, A hybrid processing system for large-scale
  traffic sensor data, IEEE Access 3 (2015) 2341--2351.

\bibitem{cite67}
M.~M. Rathore, A.~Ahmad, A.~Paul, G.~Jeon, Efficient graph-oriented smart
  transportation using internet of things generated big data, in: 2015 11th
  International Conference on Signal-Image Technology \& Internet-Based Systems
  (SITIS), IEEE, 2015, pp. 512--519.

\bibitem{cite40}
C.~Costa, M.~Y. Santos, Improving cities sustainability through the use of data
  mining in a context of big city data, in: The 2015 International Conference
  of Data Mining and Knowledge Engineering, Vol.~1, IAENG, 2015, pp. 320--325.

\bibitem{cite37}
A.~J. Jara, D.~Genoud, Y.~Bocchi, Big data in smart cities: from poisson to
  human dynamics, in: Advanced Information Networking and Applications
  Workshops (WAINA), 2014 28th International Conference on, IEEE, 2014, pp.
  785--790.

\bibitem{cite38}
H.~Wang, O.~L. Osen, G.~Li, W.~Li, H.-N. Dai, W.~Zeng, Big data and industrial
  internet of things for the maritime industry in northwestern norway, in:
  TENCON 2015-2015 IEEE Region 10 Conference, IEEE, 2015, pp. 1--5.

\bibitem{cite39}
P.~Barnaghi, M.~Bermudez-Edo, R.~T{\"o}njes, Challenges for quality of data in
  smart cities, Journal of Data and Information Quality (JDIQ) 6~(2-3) (2015)
  6.

\bibitem{cite36}
A.~Sheth, C.~Henson, S.~S. Sahoo, Semantic sensor web, IEEE Internet computing
  12~(4) (2008) 78--83.

\bibitem{cite43}
M.~A. Kafi, Y.~Challal, D.~Djenouri, M.~Doudou, A.~Bouabdallah, N.~Badache, A
  study of wireless sensor networks for urban traffic monitoring: applications
  and architectures, Procedia computer science 19 (2013) 617--626.

\bibitem{cite49}
D.~Toshniwal, et~al., Clustering techniques for streaming data-a survey, in:
  Advance Computing Conference (IACC), 2013 IEEE 3rd International, IEEE, 2013,
  pp. 951--956.

\bibitem{cite58}
V.~Jakkula, D.~Cook, Outlier detection in smart environment structured power
  datasets, in: Sixth International Conference on Intelligent Environments
  (IE), 2010, IEEE, 2010, pp. 29--33.

\bibitem{cite41}
P.~Ni, C.~Zhang, Y.~Ji, A hybrid method for short-term sensor data forecasting
  in internet of things, in: 2014 11th International Conference on Fuzzy
  Systems and Knowledge Discovery (FSKD), 2014.

\bibitem{cite55}
X.~Ma, Y.-J. Wu, Y.~Wang, F.~Chen, J.~Liu, Mining smart card data for transit
  riders’ travel patterns, Transportation Research Part C: Emerging
  Technologies 36 (2013) 1--12.

\bibitem{cite45}
W.~Derguech, E.~Bruke, E.~Curry, An autonomic approach to real-time predictive
  analytics using open data and internet of things, in: Ubiquitous Intelligence
  and Computing, 2014 IEEE 11th Intl Conf on and IEEE 11th Intl Conf on and
  Autonomic and Trusted Computing, and IEEE 14th Intl Conf on Scalable
  Computing and Communications and Its Associated Workshops (UTC-ATC-ScalCom),
  IEEE, 2014, pp. 204--211.

\bibitem{cite50}
W.~Han, Y.~Gu, Y.~Zhang, L.~Zheng, Data driven quantitative trust model for the
  internet of agricultural things, in: Internet of Things (IOT), 2014
  International Conference on the, IEEE, 2014, pp. 31--36.

\bibitem{cite46}
A.~M. Souza, J.~R. Amazonas, An outlier detect algorithm using big data
  processing and internet of things architecture, Procedia Computer Science 52
  (2015) 1010--1015.

\bibitem{cite51}
D.~N. Monekosso, P.~Remagnino, Data reconciliation in a smart home sensor
  network, Expert Systems with Applications 40~(8) (2013) 3248--3255.

\bibitem{cite47}
M.~Shukla, Y.~Kosta, P.~Chauhan, Analysis and evaluation of outlier detection
  algorithms in data streams, in: International Conference on Computer,
  Communication and Control (IC4), 2015, IEEE, 2015, pp. 1--8.

\bibitem{cite52}
A.~Shilton, S.~Rajasegarar, C.~Leckie, M.~Palaniswami, Dp1svm: A dynamic planar
  one-class support vector machine for internet of things environment, in:
  International Conference on Recent Advances in Internet of Things (RIoT),
  2015, IEEE, 2015, pp. 1--6.

\bibitem{barber2012}
D.~Barber, Bayesian reasoning and machine learning, Cambridge University Press,
  2012.

\bibitem{bishop1}
C.~M. Bishop, Pattern Recognition and Machine Learning, Springer, 2006.

\bibitem{murphy2012}
K.~P. Murphy, Machine learning: a probabilistic perspective, MIT press, 2012.

\bibitem{Goodfellow-et-al-2016}
I.~G.~Y. Bengio, A.~Courville, Deep learning, book in preparation for MIT Press
  (2016).

\bibitem{cover1967}
T.~Cover, P.~Hart, Nearest neighbor pattern classification, IEEE transactions
  on information theory 13~(1) (1967) 21--27.

\bibitem{jagadish2005}
H.~V. Jagadish, B.~C. Ooi, K.-L. Tan, C.~Yu, R.~Zhang, idistance: An adaptive
  b+-tree based indexing method for nearest neighbor search, ACM Transactions
  on Database Systems (TODS) 30~(2) (2005) 364--397.

\bibitem{mccallum1998}
A.~McCallum, K.~Nigam, et~al., A comparison of event models for naive bayes
  text classification, in: AAAI-98 workshop on learning for text
  categorization, Vol. 752, Citeseer, 1998, pp. 41--48.

\bibitem{zhang2004o}
H.~Zhang, The optimality of naive bayes, AA 1~(2) (2004) 3.

\bibitem{cortes1995s}
C.~Cortes, V.~Vapnik, Support-vector networks, Machine learning 20~(3) (1995)
  273--297.

\bibitem{guyon1993a}
I.~Guyon, B.~Boser, V.~Vapnik, Automatic capacity tuning of very large
  vc-dimension classifiers, Advances in neural information processing systems
  (1993) 147--147.

\bibitem{cristianini2000i}
N.~Cristianini, J.~Shawe-Taylor, An introduction to support vector machines and
  other kernel-based learning methods, Cambridge university press, 2000.

\bibitem{scholkopf2001l}
B.~Scholkopf, A.~J. Smola, Learning with kernels: support vector machines,
  regularization, optimization, and beyond, MIT press, 2001.

\bibitem{neter1996}
J.~Neter, M.~H. Kutner, C.~J. Nachtsheim, W.~Wasserman, Applied linear
  statistical models, Vol.~4, Irwin Chicago, 1996.

\bibitem{seber2012}
G.~A. Seber, A.~J. Lee, Linear regression analysis, Vol. 936, John Wiley \&
  Sons, 2012.

\bibitem{montgomery2015}
D.~C. Montgomery, E.~A. Peck, G.~G. Vining, Introduction to linear regression
  analysis, John Wiley \& Sons, 2015.

\bibitem{smola1997s}
A.~Smola, V.~Vapnik, Support vector regression machines, Advances in neural
  information processing systems 9 (1997) 155--161.

\bibitem{smola2004t}
A.~J. Smola, B.~Sch{\"o}lkopf, A tutorial on support vector regression,
  Statistics and computing 14~(3) (2004) 199--222.

\bibitem{breiman1984}
L.~Breiman, J.~Friedman, C.~J. Stone, R.~A. Olshen, Classification and
  regression trees, CRC press, 1984.

\bibitem{prasad2006}
A.~M. Prasad, L.~R. Iverson, A.~Liaw, Newer classification and regression tree
  techniques: bagging and random forests for ecological prediction, Ecosystems
  9~(2) (2006) 181--199.

\bibitem{loh2011}
W.-Y. Loh, Classification and regression trees, Wiley Interdisciplinary
  Reviews: Data Mining and Knowledge Discovery 1~(1) (2011) 14--23.

\bibitem{breiman2001}
L.~Breiman, Random forests, Machine learning 45~(1) (2001) 5--32.

\bibitem{breiman1996b}
L.~Breiman, Bagging predictors, Machine learning 24~(2) (1996) 123--140.

\bibitem{likas2003}
A.~Likas, N.~Vlassis, J.~J. Verbeek, The global k-means clustering algorithm,
  Pattern recognition 36~(2) (2003) 451--461.

\bibitem{Ng1}
A.~Coates, A.~Y. Ng, Learning Feature Representations with K-Means, Springer
  Berlin Heidelberg, Berlin, Heidelberg, 2012, pp. 561--580.
\newblock \href {http://dx.doi.org/10.1007/978-3-642-35289-8_30}
  {\path{doi:10.1007/978-3-642-35289-8_30}}.

\bibitem{jumutc2015}
V.~Jumutc, R.~Langone, J.~A. Suykens, Regularized and sparse stochastic k-means
  for distributed large-scale clustering, in: Big Data (Big Data), 2015 IEEE
  International Conference on, IEEE, 2015, pp. 2535--2540.

\bibitem{ester1996d}
M.~Ester, H.-P. Kriegel, J.~Sander, X.~Xu, et~al., A density-based algorithm
  for discovering clusters in large spatial databases with noise., in: Kdd,
  Vol.~96, 1996, pp. 226--231.

\bibitem{kriegel2011d}
H.-P. Kriegel, P.~Kr{\"o}ger, J.~Sander, A.~Zimek, Density-based clustering,
  Wiley Interdisciplinary Reviews: Data Mining and Knowledge Discovery 1~(3)
  (2011) 231--240.

\bibitem{campello2013d}
R.~J. Campello, D.~Moulavi, J.~Sander, Density-based clustering based on
  hierarchical density estimates, in: Pacific-Asia Conference on Knowledge
  Discovery and Data Mining, Springer, 2013, pp. 160--172.

\bibitem{pearson1901}
K.~Pearson, Liii. on lines and planes of closest fit to systems of points in
  space, The London, Edinburgh, and Dublin Philosophical Magazine and Journal
  of Science 2~(11) (1901) 559--572.

\bibitem{hotelling1933}
H.~Hotelling, Analysis of a complex of statistical variables into principal
  components., Journal of educational psychology 24~(6) (1933) 417.

\bibitem{jolliffe2002}
I.~Jolliffe, Principal component analysis, Wiley Online Library, 2002.

\bibitem{abdi2010}
H.~Abdi, L.~J. Williams, Principal component analysis, Wiley Interdisciplinary
  Reviews: Computational Statistics 2~(4) (2010) 433--459.

\bibitem{bro2014}
R.~Bro, A.~K. Smilde, Principal component analysis, Analytical Methods 6~(9)
  (2014) 2812--2831.

\bibitem{hotelling1936}
H.~Hotelling, Relations between two sets of variates, Biometrika 28~(3/4)
  (1936) 321--377.

\bibitem{bach2002}
F.~R. Bach, M.~I. Jordan, Kernel independent component analysis, Journal of
  machine learning research 3~(Jul) (2002) 1--48.

\bibitem{lecun1998}
Y.~LeCun, L.~Bottou, Y.~Bengio, P.~Haffner, Gradient-based learning applied to
  document recognition, Proceedings of the IEEE 86~(11) (1998) 2278--2324.

\bibitem{glorot2010understanding}
X.~Glorot, Y.~Bengio, Understanding the difficulty of training deep feedforward
  neural networks., in: Aistats, Vol.~9, 2010, pp. 249--256.

\bibitem{eberhart2014}
R.~C. Eberhart, Neural network PC tools: a practical guide, Academic Press,
  2014.

\bibitem{he2015}
K.~He, X.~Zhang, S.~Ren, J.~Sun, Deep residual learning for image recognition,
  arXiv preprint arXiv:1512.03385.

\bibitem{lecun2015}
Y.~LeCun, Y.~Bengio, G.~Hinton, Deep learning, Nature 521~(7553) (2015)
  436--444.

\bibitem{scholkopf2001e}
B.~Sch{\"o}lkopf, J.~C. Platt, J.~Shawe-Taylor, A.~J. Smola, R.~C. Williamson,
  Estimating the support of a high-dimensional distribution, Neural computation
  13~(7) (2001) 1443--1471.

\bibitem{ratsch2002c}
G.~Ratsch, S.~Mika, B.~Scholkopf, K.-R. Muller, Constructing boosting
  algorithms from svms: an application to one-class classification, IEEE
  Transactions on Pattern Analysis and Machine Intelligence 24~(9) (2002)
  1184--1199.

\bibitem{cite56}
C.-T. Do, A.~Douzal-Chouakria, S.~Mari{\'e}, M.~Rombaut, Multiple metric
  learning for large margin knn classification of time series, in: Signal
  Processing Conference (EUSIPCO), 2015 23rd European, IEEE, 2015, pp.
  2346--2350.

\bibitem{metsis2006}
V.~Metsis, I.~Androutsopoulos, G.~Paliouras, Spam filtering with naive
  bayes-which naive bayes?, in: CEAS, 2006, pp. 27--28.

\bibitem{webb2005}
G.~I. Webb, J.~R. Boughton, Z.~Wang, Not so naive bayes: aggregating
  one-dependence estimators, Machine learning 58~(1) (2005) 5--24.

\bibitem{gould2000}
N.~I. Gould, P.~L. Toint, A quadratic programming bibliography, Numerical
  Analysis Group Internal Report 1 (2000) 32.

\bibitem{platt1998s}
J.~Platt, et~al., Sequential minimal optimization: A fast algorithm for
  training support vector machines.

\bibitem{zhu2009p}
Z.~A. Zhu, W.~Chen, G.~Wang, C.~Zhu, Z.~Chen, P-packsvm: Parallel primal
  gradient descent kernel svm, in: 2009 Ninth IEEE International Conference on
  Data Mining, IEEE, 2009, pp. 677--686.

\bibitem{yu2009l}
C.-N.~J. Yu, T.~Joachims, Learning structural svms with latent variables, in:
  Proceedings of the 26th annual international conference on machine learning,
  ACM, 2009, pp. 1169--1176.

\bibitem{weston1999}
J.~Weston, C.~Watkins, et~al., Support vector machines for multi-class pattern
  recognition., in: ESANN, Vol.~99, 1999, pp. 219--224.

\bibitem{hsu2002c}
C.-W. Hsu, C.-J. Lin, A comparison of methods for multiclass support vector
  machines, IEEE transactions on Neural Networks 13~(2) (2002) 415--425.

\bibitem{kim2003c}
H.-C. Kim, S.~Pang, H.-M. Je, D.~Kim, S.~Y. Bang, Constructing support vector
  machine ensemble, Pattern recognition 36~(12) (2003) 2757--2767.

\bibitem{foody2004r}
G.~M. Foody, A.~Mathur, A relative evaluation of multiclass image
  classification by support vector machines, IEEE Transactions on geoscience
  and remote sensing 42~(6) (2004) 1335--1343.

\bibitem{leslie2002s}
C.~S. Leslie, E.~Eskin, W.~S. Noble, The spectrum kernel: A string kernel for
  svm protein classification., in: Pacific symposium on biocomputing, Vol.~7,
  2002, pp. 566--575.

\bibitem{poggio2001}
T.~Poggio, G.~Cauwenberghs, Incremental and decremental support vector machine
  learning, Advances in neural information processing systems 13 (2001) 409.

\bibitem{cite42}
M.~A. Khan, A.~Khan, M.~N. Khan, S.~Anwar, A novel learning method to classify
  data streams in the internet of things, in: Software Engineering Conference
  (NSEC), 2014 National, IEEE, 2014, pp. 61--66.

\bibitem{trevor2001}
H.~Trevor, T.~Robert, F.~Jerome, The elements of statistical learning: data
  mining, inference and prediction, New York: Springer-Verlag 1~(8) (2001)
  371--406.

\bibitem{caruana2006}
R.~Caruana, A.~Niculescu-Mizil, An empirical comparison of supervised learning
  algorithms, in: Proceedings of the 23rd international conference on Machine
  learning, ACM, 2006, pp. 161--168.

\bibitem{Shotton2013}
J.~Shotton, T.~Sharp, A.~Kipman, A.~Fitzgibbon, M.~Finocchio, A.~Blake,
  M.~Cook, R.~Moore, Real-time human pose recognition in parts from single
  depth images, Communications of the ACM 56~(1) (2013) 116--124.

\bibitem{au2010m}
T.~Au, M.-L.~I. Chin, G.~Ma, Mining rare events data by sampling and boosting:
  A case study, in: International Conference on Information Systems, Technology
  and Management, Springer, 2010, pp. 373--379.

\bibitem{sahu2011i}
A.~Sahu, G.~Runger, D.~Apley, Image denoising with a multi-phase kernel
  principal component approach and an ensemble version, in: 2011 IEEE Applied
  Imagery Pattern Recognition Workshop (AIPR), IEEE, 2011, pp. 1--7.

\bibitem{shinde2014}
A.~Shinde, A.~Sahu, D.~Apley, G.~Runger, Preimages for variation patterns from
  kernel pca and bagging, IIE Transactions 46~(5) (2014) 429--456.

\bibitem{cite48}
X.~Tao, C.~Ji, Clustering massive small data for iot, in: 2nd International
  Conference on Systems and Informatics (ICSAI), 2014, IEEE, 2014, pp.
  974--978.

\bibitem{cite59}
H.~Hromic, D.~Le~Phuoc, M.~Serrano, A.~Antoni{\'c}, I.~P. {\v{Z}}arko,
  C.~Hayes, S.~Decker, Real time analysis of sensor data for the internet of
  things by means of clustering and event processing, in: 2015 IEEE
  International Conference on Communications (ICC), IEEE, 2015, pp. 685--691.

\bibitem{ccelik2011an}
M.~{\c{C}}elik, F.~Dada{\c{s}}er-{\c{C}}elik, A.~{\c{S}}. Dokuz, Anomaly
  detection in temperature data using dbscan algorithm, in: Innovations in
  Intelligent Systems and Applications (INISTA), 2011 International Symposium
  on, IEEE, 2011, pp. 91--95.

\bibitem{golub2012matrix}
G.~H. Golub, C.~F. Van~Loan, Matrix computations, Vol.~3, JHU Press, 2012.

\bibitem{sirovich1987turbulence}
L.~Sirovich, Turbulence and the dynamics of coherent structures part i:
  coherent structures, Quarterly of applied mathematics 45~(3) (1987) 561--571.

\bibitem{cybenko1989}
G.~Cybenko, Approximation by superpositions of a sigmoidal function,
  Mathematics of control, signals and systems 2~(4) (1989) 303--314.

\bibitem{hornik1991}
K.~Hornik, Approximation capabilities of multilayer feedforward networks,
  Neural networks 4~(2) (1991) 251--257.

\bibitem{fukushima1980}
K.~Fukushima, Neocognitron: A self-organizing neural network model for a
  mechanism of pattern recognition unaffected by shift in position, Biological
  cybernetics 36~(4) (1980) 193--202.

\bibitem{blum1992teaching}
W.~BLUM, D.~BURGHES, N.~GREEN, G.~KAISER-MESSMER, Teaching and learning of
  mathematics and its applications: first results from a comparative empirical
  study in england and germany, Teaching Mathematics and its Applications
  11~(3) (1992) 112--123.

\bibitem{schmidhuber2015}
J.~Schmidhuber, Deep learning in neural networks: An overview, Neural Networks
  61 (2015) 85--117.

\bibitem{cite57}
I.~Kotenko, I.~Saenko, F.~Skorik, S.~Bushuev, Neural network approach to
  forecast the state of the internet of things elements, in: XVIII
  International Conference on Soft Computing and Measurements (SCM), 2015,
  IEEE, 2015, pp. 133--135.

\bibitem{dietterich2002m}
T.~G. Dietterich, Machine learning for sequential data: A review, in: Joint
  IAPR International Workshops on Statistical Techniques in Pattern Recognition
  (SPR) and Structural and Syntactic Pattern Recognition (SSPR), Springer,
  2002, pp. 15--30.

\bibitem{baum1967i}
L.~E. Baum, J.~A. Eagon, et~al., An inequality with applications to statistical
  estimation for probabilistic functions of markov processes and to a model for
  ecology, Bull. Amer. Math. Soc 73~(3) (1967) 360--363.

\bibitem{rabiner1989t}
L.~R. Rabiner, A tutorial on hidden markov models and selected applications in
  speech recognition, Proceedings of the IEEE 77~(2) (1989) 257--286.

\bibitem{sejnowski1987p}
T.~J. Sejnowski, C.~R. Rosenberg, Parallel networks that learn to pronounce
  english text, Complex systems 1~(1) (1987) 145--168.

\bibitem{kalman1961n}
R.~E. Kalman, R.~S. Bucy, New results in linear filtering and prediction
  theory, Journal of basic engineering 83~(1) (1961) 95--108.

\bibitem{lafferty2001c}
J.~Lafferty, A.~McCallum, F.~Pereira, Conditional random fields: Probabilistic
  models for segmenting and labeling sequence data, in: Proceedings of the
  eighteenth international conference on machine learning, ICML, Vol.~1, 2001,
  pp. 282--289.

\bibitem{williams1989l}
R.~J. Williams, D.~Zipser, A learning algorithm for continually running fully
  recurrent neural networks, Neural computation 1~(2) (1989) 270--280.

\bibitem{mccallum2000m}
A.~McCallum, D.~Freitag, F.~C. Pereira, Maximum entropy markov models for
  information extraction and segmentation., in: Icml, Vol.~17, 2000, pp.
  591--598.

\bibitem{sak2014l}
H.~Sak, A.~W. Senior, F.~Beaufays, Long short-term memory recurrent neural
  network architectures for large scale acoustic modeling., in: INTERSPEECH,
  2014, pp. 338--342.

\bibitem{graves2009n}
A.~Graves, M.~Liwicki, S.~Fern{\'a}ndez, R.~Bertolami, H.~Bunke,
  J.~Schmidhuber, A novel connectionist system for unconstrained handwriting
  recognition, IEEE transactions on pattern analysis and machine intelligence
  31~(5) (2009) 855--868.

\bibitem{pardo2005m}
B.~Pardo, W.~Birmingham, Modeling form for on-line following of musical
  performances, in: PROCEEDINGS OF THE NATIONAL CONFERENCE ON ARTIFICIAL
  INTELLIGENCE, Vol.~20, Menlo Park, CA; Cambridge, MA; London; AAAI Press; MIT
  Press; 1999, 2005, p. 1018.

\bibitem{hodge2004s}
V.~J. Hodge, J.~Austin, A survey of outlier detection methodologies, Artificial
  intelligence review 22~(2) (2004) 85--126.

\bibitem{chandola2009a}
V.~Chandola, A.~Banerjee, V.~Kumar, Anomaly detection: A survey, ACM computing
  surveys (CSUR) 41~(3) (2009) 15.

\bibitem{milacski2015r}
Z.~{\'A}. Milacski, M.~Ludersdorfer, A.~L{\H{o}}rincz, P.~van~der Smagt, Robust
  detection of anomalies via sparse methods, in: International Conference on
  Neural Information Processing, Springer, 2015, pp. 419--426.

\bibitem{denning1987i}
D.~E. Denning, An intrusion-detection model, IEEE Transactions on software
  engineering~(2) (1987) 222--232.

\bibitem{tomek1976e}
I.~Tomek, An experiment with the edited nearest-neighbor rule, IEEE
  Transactions on systems, Man, and Cybernetics~(6) (1976) 448--452.

\bibitem{smith2011i}
M.~R. Smith, T.~Martinez, Improving classification accuracy by identifying and
  removing instances that should be misclassified, in: Neural Networks (IJCNN),
  The 2011 International Joint Conference on, IEEE, 2011, pp. 2690--2697.

\bibitem{rajasegarar2010c}
S.~Rajasegarar, C.~Leckie, J.~C. Bezdek, M.~Palaniswami, Centered
  hyperspherical and hyperellipsoidal one-class support vector machines for
  anomaly detection in sensor networks, IEEE Transactions on Information
  Forensics and Security 5~(3) (2010) 518--533.

\bibitem{heller2003o}
K.~A. Heller, K.~M. Svore, A.~D. Keromytis, S.~J. Stolfo, One class support
  vector machines for detecting anomalous windows registry accesses, in: Proc.
  of the workshop on Data Mining for Computer Security, Vol.~9, 2003.

\bibitem{zhang2015an}
M.~Zhang, B.~Xu, J.~Gong, An anomaly detection model based on one-class svm to
  detect network intrusions, in: 2015 11th International Conference on Mobile
  Ad-hoc and Sensor Networks (MSN), IEEE, 2015, pp. 102--107.

\bibitem{zhang2009ad}
Y.~Zhang, N.~Meratnia, P.~Havinga, Adaptive and online one-class support vector
  machine-based outlier detection techniques for wireless sensor networks, in:
  Advanced Information Networking and Applications Workshops, 2009. WAINA'09.
  International Conference on, IEEE, 2009, pp. 990--995.

\bibitem{cite60}
S.~Hu, Research on data fusion of the internet of things, in: Logistics,
  Informatics and Service Sciences (LISS), 2015 International Conference on,
  IEEE, 2015, pp. 1--5.

\end{thebibliography}








\end{document}